\documentclass{article}

\usepackage[nonatbib,preprint]{neurips_data_2024}
\usepackage[square,numbers]{natbib}

\usepackage[utf8]{inputenc} 
\usepackage[T1]{fontenc}    
\usepackage{hyperref}       
\usepackage{url}            
\usepackage{booktabs}       
\usepackage{amsfonts}       
\usepackage{nicefrac}       
\usepackage{microtype}      
\usepackage{xcolor}         
\usepackage{listings}       
\usepackage[many]{tcolorbox} 
\usepackage{tikz}           
\usepackage{xspace}
\usepackage{longtable}      
\usepackage{subcaption}     
\usepackage{dcolumn} 

\usepackage{multirow}
\usepackage{colortbl}
\usepackage[textwidth=3cm]{todonotes}
\usepackage{color}

\def\myComments{0}

\newcommand{\ourmethod}{\textsc{APEx}\xspace}
\newcommand{\ourmethodFull}{\textsc{Automatic Programming of Experiments}\xspace}
\newcommand{\prompt}[1]{{\texttt{\color{codegray} {#1}}}}

\newcommand{\inlineColorbox}[2]{\begingroup\setlength{\fboxsep}{1pt}\colorbox{#1}{\hspace*{2pt}\vphantom{Ay}#2\hspace*{2pt}}\endgroup}

\definecolor{DrawioBlue}{RGB}{218,232,252}
\definecolor{DrawioOrange}{RGB}{255,230,204}
\definecolor{DrawioGreen1}{RGB}{204,255,204}
\definecolor{DrawioGreen}{RGB}{213,232,212}
\definecolor{DrawioPurple}{RGB}{225,213,231}
\definecolor{DrawioOrchestrator}{RGB}{217,234,211}
\definecolor{DrawioEngine}{RGB}{234,153,153}
\definecolor{DrawioGenerator}{RGB}{230,145,56}
\definecolor{DrawioLibrary}{RGB}{180,167,214}
\definecolor{DrawioQuery}{RGB}{213,166,189}
\definecolor{DrawioReport}{RGB}{153,153,153}
\definecolor{DrawioExperiment}{RGB}{255, 229, 153}
\definecolor{DrawioRed}{RGB}{248,206,204}
\definecolor{DrawioYellow}{RGB}{255, 242, 207}
\definecolor{lightgray}{RGB}{225,225,225}

\newcommand{\eg}{\textit{e}.\textit{g}.}
\newcommand{\ie}{\textit{i}.\textit{e}.}

\ifthenelse{\equal{\myComments}{1}}
{
\newcommand{\lcm}[3]{{\color{#2} {[#1] #3}}}}
{\newcommand{\lcm}[3]{}
}

\ifthenelse{\equal{\myComments}{1}}
{
\newcommand{\scm}[3]{{\todo[color=#2!40]{\textbf{\footnotesize[#1]} #3}}}}
{\newcommand{\scm}[3]{}
}

\definecolor{DrawioBlue}{RGB}{218,232,252}
\definecolor{DrawioOrange}{RGB}{255,230,204}
\definecolor{DrawioGreen1}{RGB}{204,255,204}
\definecolor{DrawioGreen}{RGB}{213,232,212}
\definecolor{DrawioPurple}{RGB}{225,213,231}
\definecolor{DrawioRed}{RGB}{248,206,204}
\definecolor{DrawioYellow}{RGB}{255, 242, 207}

\definecolor{codegreen}{rgb}{0,0.6,0}
\definecolor{codegray}{rgb}{0.5,0.5,0.5}
\definecolor{codepurple}{rgb}{0.58,0,0.82}
\definecolor{backcolour}{rgb}{0.95,0.95,0.92}
\definecolor{arylideyellow}{rgb}{0.91, 0.84, 0.42}
\definecolor{beaublue}{rgb}{0.74, 0.83, 0.9}

\lstdefinestyle{mystyle}{
    backgroundcolor=\color{backcolour},   
    commentstyle=\color{codegreen},
    keywordstyle=\color{magenta},
    numberstyle=\tiny\color{codegray},
    stringstyle=\color{codepurple},
    basicstyle=\ttfamily\scriptsize,
    breakatwhitespace=false,         
    breaklines=true,                 
    captionpos=b,                    
    keepspaces=true,                 
    numbers=left,                    
    numbersep=5pt,                  
    showspaces=false,                
    showstringspaces=false,
    showtabs=false,                  
    tabsize=2
}

\title{Automatic benchmarking of large multimodal models\\via iterative experiment programming}

\author{%
  Alessandro Conti\textsuperscript{1,}\thanks{Currently at Apple. Correspondence to: \texttt{alessandro.conti-1@unitn.it}.}\quad
  Enrico Fini\textsuperscript{1,*}\quad
  Paolo Rota\textsuperscript{1}\quad \vspace{.1em}\\
  \textbf{Yiming Wang\textsuperscript{2}}\quad
  \textbf{Massimiliano Mancini\textsuperscript{1}} 
  \textbf{Elisa Ricci\textsuperscript{1,2}} \vspace{1em} \\
  \textsuperscript{1}University of Trento \vspace{.1em}\\ \textsuperscript{2}Fondazione Bruno Kessler 
}

\begin{document}

\maketitle

\begin{abstract}
Assessing the capabilities of large multimodal models (LMMs) often requires the creation of ad-hoc evaluations.
Currently, building new benchmarks requires tremendous amounts of manual work for each specific analysis.
This makes the evaluation process tedious and costly.
In this paper, we present \ourmethod, \ourmethodFull, the first framework for automatic benchmarking of LMMs.
Given a research question expressed in natural language, \ourmethod leverages a large language model (LLM) and a library of pre-specified tools to generate a set of experiments for the model at hand, and progressively compile a scientific report.
The report drives the testing procedure: based on the current status of the investigation, \ourmethod chooses which experiments to perform and whether the results are sufficient to draw conclusions.
Finally, the LLM refines the report, presenting the results to the user in natural language.
Thanks to its modularity, our framework is flexible and extensible as new tools become available.
Empirically, \ourmethod reproduces the findings of existing studies while allowing for arbitrary analyses and hypothesis testing.
{Code is available at \href{https://github.com/altndrr/apex}{https://github.com/altndrr/apex}}.
\end{abstract}

\section{Introduction}
The more powerful machine learning models become, the greater the community's interest in testing their capabilities and limitations.
Large Multimodal Models (LMMs)\cite{li2023blip,laurenccon2024obelics} have been investigated by several studies with extensive analysis of their strengths and weaknesses~\cite{udandarao2023visual,yuksekgonul2022bow,hsieh2024sugarcrepe}.
Recent works~\cite{chen2024benchmarking,zhao2024evaluating} typically adhere to a common workflow: (i) selecting models for testing, (ii) devising specialized benchmarks tailored to address specific research questions (which may entail data collection and annotation), and (iii) evaluating the models on these benchmarks, analyzing the results, and drawing conclusions.

Let us examine a simple example.
Consider the question: \prompt{Do LMMs understand colors?} Answering this question requires testing one or multiple models on color recognition tasks.
To perform this experimental evaluation, various considerations can arise.
How many colors should be included in the test?
How complex should the color palette be?
If a model shows proficiency with a specific set of colors, should we conclude the test or continue with further examination?
In other words, performing such experimental validation requires the researchers to perform a great deal of manual work: designing the benchmark, testing the models, analyzing the results, and determining whether the depth of the analyses is sufficient to draw conclusions.
This process is not only tedious but also complex, particularly in designing benchmarks, as it requires significant domain expertise.
Additionally, data collection and curation can be notoriously expensive, and the protocol design and result analyses can be bound to subjectivity.
These considerations lead us to investigate a framework capable of autonomously designing and executing a wide range of experiments aimed at evaluating the capabilities of existing LMMs.

In this paper, we propose to address this question following the principles of programming visual tools via Large Language Models (LLMs)~\cite{touvron2023llama}.
Our framework, named \ourmethodFull (\ourmethod), redacts a scientific report to answer any user-specified inquiry by first instantiating a structured empty report and then iteratively improving it through a series of experiments.
Specifically, an LLM processes the user request and: (i) generates a suitable benchmark via image retrieval, generation, or augmentation; (ii) deploys models to test; (iii) conducts the experiment; (iv) collects and analyzes the results.
The findings are incorporated into the report, which is then fed back to the LLM to determine whether the information is sufficient to answer the initial inquiry.
If not, \ourmethod creates a new benchmark and runs more experiments.
This iterative process continues until the LLM judges the report comprehensive enough to address the user's question.
We evaluate \ourmethod with a wide range of queries to demonstrate its flexibility in designing benchmarks and automatically conducting experiments, reaching conclusions that are valid and insightful.
To demonstrate that \ourmethod actually produces reliable reports, we show that it can successfully lead to the core findings of existing manually engineered benchmark studies~\cite{udandarao2023visual}.
Moreover, \ourmethod can flexibly handle queries of various granularity, allowing users to thoroughly explore the strengths and weaknesses of existing LMMs.
Finally, the modularity of \ourmethod enables easy inclusion of new tools and capabilities in the future, widening its application scope.

To summarize, this work makes the following contributions:
\begin{itemize}
    \item We introduce \ourmethod, the first automated benchmark framework to test various capabilities of LMMs at user request;
    \item \ourmethod automates benchmark design, experiments execution, and results analysis.
    Its modular design makes it extensible to incorporate other tools and functionalities;
    \item We show that \ourmethod outputs valid and comprehensive reports, being able to reproduce findings of previous studies.
    \item \ourmethod facilitates the discovery of novel strengths and weaknesses of existing LMMs by addressing arbitrary queries at different granularity.
\end{itemize}

\section{Related work}
Our work is related to previous research on modular machine learning and studies that investigate the capabilities of LMMs.
Below, we review the most relevant works.

\textbf{Visual programming.} Composing atomic operations to realize complex functions is a popular approach in Visual Question Answering (VQA)~\cite{antol2015vqa}.
In this context, modular neural networks~\cite{andreas2016neural,subramanian2023modular} decompose questions~(\eg, \prompt{How many dogs are in the image?}) into sub-tasks (\eg, detecting, counting) that can solve the whole query when their execution is chained (\eg, {counting} after {detection}).
This paradigm has recently evolved with the enhanced reasoning capabilities of LLMs~\cite{openai2022chatgpt}.
Specifically, in Natural Language Processing, emerged frameworks augment LLMs with the ability to use external tools~\cite{schick2024toolformer,parisi2022talm,qin2023toolllm}.
This is usually achieved by specifying the set of tools within the prompt (\eg, as APIs) and providing examples of use cases, performing in-context learning \cite{brown2020language}.
For vision, VisProg~\cite{gupta2023visprog} and ViperGPT~\cite{suris2023vipergpt} proposed to use the same paradigm to solve various computer vision tasks.
Specifically, they leverage the LLMs to generate programs executed by a suite of pre-trained vision models.
Note that the LLM acts as the parser and sub-task decomposer in ~\cite{andreas2016neural}.
Subsequent works focused on two main directions.
The first is to improve the visual programming pipeline by, \eg, expanding the set of available tools~\cite{shen2024hugginggpt}, adding verification steps~\cite{lu2024chameleon}, or self-training~\cite{khan2024self}.
The second direction focuses on applying them for specific tasks, \eg, animal behavior understanding~\cite{ye2024amadeusgpt}, open-vocabulary 3D grounding~\cite{yuan2023visual}, and text-to-image generation~\cite{cho2024visual}.
Concurrently,~\cite{shaham2024maia} used a set of tools to automatically interpret neurons of deep networks.

\ourmethod~builds on previous works tackling visual tasks via LLM-based program generation~\cite{gupta2023visprog,suris2023vipergpt}.
However, in contrast to previous studies, we propose a paradigm shift introducing a novel framework for generating programs for rigorously testing LMMs, rather than solving specific tasks.

\textbf{Analyzing vision language models.} 
 Several previous studies investigated the capabilities and shortcomings of Vision Language Models (VLMs).
 For instance, \cite{thrush2022winoground,ma2023crepe} studied VLMs on fine-grained differences in textual inputs, showing issues for capturing changes in relationships, objects, and attributes.
 In \cite{yuksekgonul2022bow} the authors further analyzed the impact of words and patch orders, showing that VLMs act as bag-of-words if fine-grained relationships are not considered when forming training batches.
 Other works studied VLMs biases concepts seen in the training set.
 Examples are \cite{tang2023lemons} that shows biases in how VLMs associate attributes to objects, \cite{gavrikov2024shape} that studies shape vs texture bias, and \cite{udandarao2024no,fang2022data,mayilvahanan2023does} that show how frequency of concepts impact zero-shot generalization.
 Other works focused on testing VLMs for specific tasks such as time/location reasoning~\cite{zhang2024can}, egocentric vision~\cite{cheng2023can}, information retrieval~\cite{chen2023can}, spatial reasoning~\cite{kamath2023up}, or general multimodal understanding~\cite{chen2023towards,shi2023chef}.
 Similarly, studies explored the robustness of VLMs to distribution-shifts~\cite{qiu2023benchmarking}, corruptions~\cite{chen2024benchmarking}, adversarial attacks~\cite{zhao2024evaluating}, and missing modalities~\cite{ma2022multimodal}.
 It is important to note that most of these works required designing specific benchmarks and it is not uncommon that follow-up works identify and fix benchmark issues of previous work~\cite{hsieh2024sugarcrepe}.
 
 Differently from these works, we do not focus on a single analysis but provide a general tool for automatically creating benchmarks and inspecting what VLMs (or LMMs), can do.
 As our software is publicly available, researchers and AI professionals can not only use it for LMMs testing but also contribute to \ourmethod, adding modules for expanding its capabilities.
 
\section{\ourmethod: \ourmethodFull}
\label{sec:method}
In this section, we present our \ourmethod framework that automatically conducts experiments on different LMMs to answer a given user inquiry.
We refer to $\operatorname{query}$ as the text prompt in a question format that contains a hypothesis the user seeks to test.
The goal of \ourmethod is to provide an answer $a$ in natural language that addresses the user query through experimental evidence.

\begin{figure}
  \centering
  \includegraphics[width=0.95\linewidth]{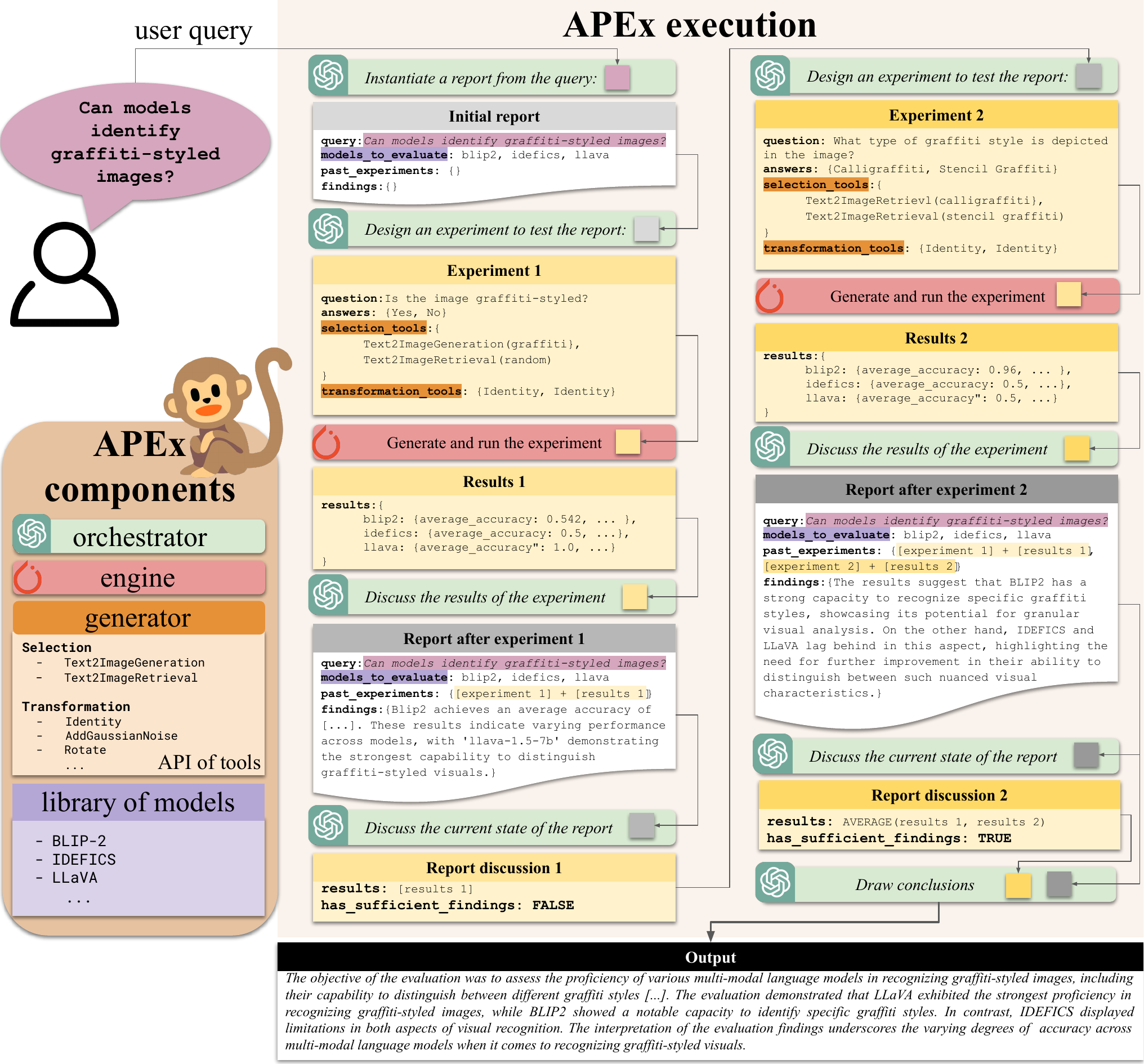}
  \caption{\textbf{\ourmethod.} Our automatic benchmarking tool has four components: an \inlineColorbox{DrawioOrchestrator}{$\operatorname{orchestrator}$} for reasoning, an \inlineColorbox{DrawioEngine}{$\operatorname{engine}$} for function execution, a benchmark \inlineColorbox{DrawioGenerator}{$\operatorname{generator}$} containing image selection and manipulation tools, and a \inlineColorbox{DrawioLibrary}{$\operatorname{library}$} of LMMs.
  Given a user \inlineColorbox{DrawioQuery}{$\operatorname{query}$}, the $\operatorname{orchestrator}$ instantiates a \inlineColorbox{DrawioReport}{$\operatorname{report}$} containing the query and the LMMs to be tested.
  Then $\operatorname{orchestrator}$ receives the $\operatorname{report}$ and specifies a first \inlineColorbox{DrawioExperiment}{$\operatorname{experiment}$} to be executed.
  The relative benchmark is generated and executed by the $\operatorname{engine}$, with the $\operatorname{results}$ collected, discussed by the $\operatorname{orchestrator}$ and added to the $\operatorname{report}$.
  The $\operatorname{orchestrator}$ repeats the experimentation loop until it is deemed sufficient to answer the query.
  In that case, the $\operatorname{orchestrator}$ summarizes the report, returning its findings.}
  \label{fig:teaser}
\end{figure}

\ourmethod is a modular system containing multiple components supported by a set of tools, as shown in Fig.~\ref{fig:teaser}.
At its core, there is the $\operatorname{orchestrator}$ module, enabled by an LLM that reasons over the available tool-set to define experiments and analyze results.
The $\operatorname{generator}$ module is equipped with an API ($\operatorname{tools}$) that can perform image retrieval, generation, and transformation to prepare the data for executing benchmark experiments.
The $\operatorname{library}$ module contains a list of pre-trained LMMs, and finally, the $\operatorname{engine}$ module conducts the experiment with the compute resource, possibly a GPU.
Note that the modular structure of \ourmethod is independent of the particular choices of each sub-module.
Thus, its capabilities can be extended as more powerful LMMs, LLMs, generators and tools become available.
For example, given a user query (\eg, \prompt{Is BLIP-2 robust to left rotations?}), \ourmethod selects the appropriate tools by compiling an experiment configuration, runs the experiment, and analyzes the results to draw conclusions.
The experiments are conducted progressively and the evidence is consolidated into a \textit{report} that holds the status of the experimental campaign.
The automatic benchmarking session terminates when the produced results are sufficiently conclusive to answer the initial query.
In the following, we detail the components of \ourmethod.

\subsection{Initialization}
\label{sec:init}

Given a $\operatorname{query}$, \ourmethod is initialized providing specific definitions for the $\operatorname{library}$ of testable models, for the benchmark $\operatorname{generator}$ tools, and for the LLM used in the $\operatorname{orchestrator}$.

When receiving an input query $\operatorname{query}$ from the user, the first step of \ourmethod is to instantiate an empty $\operatorname{report}$ as a JSON formatted document.
The $\operatorname{report}$ will contain a snapshot of the request, the performed experiments, and their results.
In practice, it serves as a read-write buffer for the $\operatorname{orchestrator}$, driving the decisions regarding what to test and when to halt.
We initialize the report by feeding the $\operatorname{query}$ to the $\operatorname{orchestrator}$, \ie, $\operatorname{report} = \operatorname{orchestrator} \left( \left[\operatorname{prompt}_S,\operatorname{prompt}_I, \operatorname{query}\right] \right)$, with $\left[\cdot\right]$ being the concatenation operation, $\operatorname{prompt}_I$ being the specific initialization prompt, and $\operatorname{prompt}_S$ the system prompt containing the task specification, the models' $\operatorname{library}$, and the benchmark $\operatorname{generator}$ tools.
Both  $\operatorname{library}$ and the $\operatorname{generator}$ are specified with their documentation in the form of docstrings.
Note that we do not use in-context learning~\cite{brown2020language} (\ie, providing few program demonstrations as in~\cite{gupta2023visprog,suris2023vipergpt}), but directly use the docstrings as input.
This removes the need for providing good quality demonstrations, being scalable w.r.t. the number of tools~\cite{qin2023toolllm}.

The initialized $\operatorname{report}$ contains only the specific $\operatorname{query}$ and the set of LMMs to test (\ie, \prompt{\detokenize{models_to_evaluate}} in Fig.~\ref{fig:teaser}).
These models will be the ones used in the following iterative experimentation.
We may target all models for generic queries (\eg, \prompt{Can models...?}) or specific models for more detailed ones (\eg, BLIP-2 if the query is \prompt{Can BLIP-2...?}).

\subsection{Iterative Experimentation}
\label{sec:iter}
The core of \ourmethod is the experimentation loop.
In this loop, the $\operatorname{orchestrator}$ takes as input the query, the history of experiments and results (\ie, the $\operatorname{report}$), and formulates new experiments if needed.
In particular, it considers the question that should be answered with the current experimental results and then generates a suitable benchmark, testing the selected models on it.
Specifically, we define the experiment by feeding the LLM with the $\operatorname{report}$ and the user query, \ie, $\operatorname{experiment} = \operatorname{orchestrator} \left( \left[\operatorname{prompt}_S,\operatorname{prompt}_E, \operatorname{report} \right] \right)$, with $\operatorname{prompt}_E$ being the specific prompt.
The $\operatorname{experiment}$ contains two parts: the goal definition and the benchmark dataset generation.
In the following, we describe each part in detail as well as the experiment execution.

\textbf{Goal definition.} 
This part defines the goal of the experiment.
It is composed of two elements: a $\operatorname{question}$ we want the LMM to answer (\eg, \prompt{Is the image blue?}, not to be confused with the $\operatorname{query}$) and the set of possible answer $\operatorname{choices}$ (\eg, "\{\prompt{yes}, \prompt{no}\}").
Note that all experiments are structured in a VQA format, where the LMM receives the question and the input image $x$.
We measure the performance using answer ranking~\cite{lin2022truthfulqa} for evaluating LMMs with multiple-choice questions (\ie, computing the likelihood that the model generates a specific answer among the list).
We allow the LMMs to abstain from answering by adding \prompt{Unknown} as an option for the answer.

\textbf{Benchmark dataset generation.} Given an $\operatorname{experiment}$, \ourmethod can generate a dataset $\mathcal{D}$ for testing.
Specifically, $\mathcal{D}$ should be formed with triplets $\mathcal{D}=\{(x_i, \operatorname{question}, \operatorname{choice}_i)\}_{i=1}^N$ where $N$ is the size of $\mathcal{D}$, $\operatorname{choice}_i$ is the ground-truth answer for the given pair $x_i$, $\operatorname{question}$.
Within $\operatorname{experiment}$ we have the specific $\operatorname{tools}_E$ needed to generate the dataset.

Since the ground-truth label is a function of the tools used, we associate a set of selection and transformation tools to each answer in $\operatorname{choice}$.
This is a consequence of the structured JSON output, allowing for easy matching between the elements of each field.
Note that the current set of $\operatorname{tools}_E$ comprises both image collection and generation instructions (\eg, retrieval from existing datasets~\cite{russakovsky2015imagenet}, Stable Diffusion~\cite{rombach2022high}) and the transformations to be applied (\eg, rotation, flip).
Note that the transformations are represented by Python code that the model can directly execute by outputting an interpretable function call, \eg, \prompt{src.tools.transform.OverlayColor([255, 0, 0])}.
When the $\operatorname{engine}$ produces the dataset, we evaluate these calls to augment or generate the data.

\textbf{Experiment execution.} 
\label{sec:loop}
After the benchmark generation, all the components needed to perform the experiments are available.
Specifically, the $\operatorname{engine}$ tests each of the LMMs $\operatorname{models}$ on $\mathcal{D}$, evaluating their performance.
The output of this $\operatorname{engine}$ execution is a new set of $\operatorname{results}_E$.

\subsection{Reporting and conclusion}
\label{sec:reporting}

After each experimental loop, we obtain a new result set of  $\operatorname{results}_E$.
\ourmethod expands the results and asks the $\operatorname{orchestrator}$ to discuss them, obtaining the findings $\operatorname{findings}_E=\operatorname{orchestrator}\left(\left[ \operatorname{prompt}_F, \operatorname{experiment}, \operatorname{results} \right]\right)$, where $\operatorname{prompt}_F$ is the specific prompt for extracting the findings.
The findings, in natural language, help the $\operatorname{orchestrator}$ in performing reasoning and draw conclusions beyond the quantitative metrics in $\operatorname{results}_E$.

Both the results and the findings are appended at the end of the current $\operatorname{report}$, updating its status \ie, $\operatorname{report}\gets \left[\operatorname{report}, \operatorname{results}_E, \operatorname{findings}_E \right]$.
The updated $\operatorname{report}$ contains all the evidence collected so far to answer the user query.
$\operatorname{orchestrator}$ then judges whether the updated report is sufficient to answer the user query, filling a boolean variable $\operatorname{sufficiency}=\operatorname{orchestrator}\left(\left[ \operatorname{prompt}_B, \operatorname{report}\right]\right)$, with $\operatorname{prompt}_B$ being the prompt for obtaining the boolean value.
In case the $\operatorname{orchestrator}$ deems the $\operatorname{report}$ \textit{not} sufficient to answer the $\operatorname{query}$, the $\operatorname{report}$ is fed back to the $\operatorname{orchestrator}$ to continue the experimental loop, enriching the $\operatorname{report}$ with the results of a new $\operatorname{experiment}$.

Once the $\operatorname{orchestrator}$ deems the $\operatorname{report}$ sufficient to answer the $\operatorname{query}$, \ourmethod will use the $\operatorname{orchestrator}$ to analyze the report and draw the final conclusions of the experiment, \ie, $\operatorname{conclusions}=\operatorname{orchestrator}\left(\left[\operatorname{prompt}_C, \operatorname{report}\right]\right)$, with $\operatorname{prompt}_C$ the specific prompt.
Note that $\operatorname{conclusions}$ can be customized based on the particular user needs, \eg, they can be succinct, reporting only the answer to the $\operatorname{query}$, or expanded with a summary of the experiments, their results, and possible directions.
Once $\operatorname{conclusions}$ are drawn, \ourmethod provides them as output to the user.

\subsection{Implementation details}
\label{sec:implementation details}

\textbf{{Orchestrator}.} The core component of \ourmethod is the $\operatorname{orchestrator}$, implemented as an LLM.
We used gpt-3.5-turbo~\cite{openai2022chatgpt} for its efficiency and widespread use.
To instruct the model to produce structured outputs, we use \texttt{function calling}\footnote{\url{https://platform.openai.com/docs/guides/function-calling}}, meaning the LLM outputs adhere to the structured JSON format rather than being unrestricted text, ensuring easy mapping of the model outputs to functions/tools available in \ourmethod.
In cases where the $\operatorname{orchestrator}$'s output is invalid, \eg, it lacks JSON fields, or it requests non-available tools, \ourmethod informs the $\operatorname{orchestrator}$ of the exception message (within the prompt), using this information as prior to \textit{self-heal} the previous output.
In the case of complex healing (\eg, multiple sequential failures), a new experiment is designed from scratch.
To avoid infinite loops, we set the maximum number of self-healing retries to 3 before regenerating the experiment from scratch.
Moreover, in all of our evaluations, we limit the maximum number of experiments the $\operatorname{orchestrator}$ can design to 5.

\textbf{Benchmark generation.} As stated in Sec.\ref{sec:iter} the data benchmark generation step in \ourmethod comprises both image generation tools and tools for retrieval.
For generation, we use Stable Diffusion XL Turbo \cite{podell2023sdxl} for its efficiency.
For retrieval, we use the ImageNet dataset~\cite{russakovsky2015imagenet} due to the large number of classes.
In both cases, in \ourmethod, it is possible to define an image type (\eg, photo, oil painting) and a class name (\eg, Siamese cat, sofa).
The prompts corresponding to the two available tools are \prompt{TextToImageGeneration} and \prompt{TextToImageRetrieval}, passed to \ourmethod in form of docstrings (see, \eg, Fig.~\ref{fig:teaser}, Experiment box).
To perform experiments with broader categories, \ourmethod uses the metaclasses from ImageNet-X~\cite{idrissi2022imagenet}, clustering the 1000 classes into 17 groups (\eg, plants, fruits).
When retrieval is impossible, \eg, the requested class is not available in the dataset, \ourmethod automatically resorts to generation.
We also use the special keyword \prompt{random} to denote sampling a semantic class in the dataset.
This is used whenever the experiment does not require any specific class, and all of them can be considered for the benchmark.
Regarding the transformation tools, \ie, tools to manipulate the input samples for visual variations, we follow the design as in~\cite{udandarao2023visual}.
These transformations can be roughly split into geometric, pixel, semantic, and style transforms.
Additional details are provided in the Appendix.

\textbf{LMMs evaluation.} 
While in principle \ourmethod supports a large variety of LMMs, in our implementation we demonstrate its capabilities using three widely used auto-regressively trained VLMs (referred as LMMs in~\cite{udandarao2023visual}): BLIP-2~\cite{li2023blip}, IDEFICS~\cite{laurenccon2024obelics}, and LLaVA~\cite{li2024llava}.
All the models are downloaded from HuggingFace\footnote{\url{https://huggingface.co}} and are extended to support answer ranking.
We focus on small models, using \texttt{blip2-opt-2.7b}, \texttt{idefics-9b-instruct}, and \texttt{llava-1.5-7b}.
We quantize all the models to 8 bits and use a single NVIDIA RTX A4000 for the experiments.
About the quantitative metrics in $\operatorname{results}_E$, we evaluate the methods with average accuracy and class-wise accuracy.
In cases when the models abstain from answering, we compute the abstention rate.
The quantitative metrics are reported in a key-value format organized by evaluated models (see Fig.~\ref{fig:teaser}, Results box).

\section{Experiments}
\label{sec:exp}

Our tool generates free-form text output, summarizing findings related to the input query.
An illustrative example of qualitative results obtained using \ourmethod is depicted in Fig.~\ref{fig:teaser}.
In addition, we quantitatively evaluate our framework as described below.
As an important feature is the possibility to reproduce conclusions obtained by manually designed benchmark, we first demonstrate that \ourmethod can reproduce the core conclusions of~\cite{udandarao2023visual} regarding LMMs on data type identification, \ie, on recognizing image alterations (\eg, \prompt{Can models identify vertical flip/Gaussian noise/cartoon-styled images?}).
Second, we showcase the flexibility of \ourmethod in handling queries of different nature and with different granularity.
In particular, we present the analyses of \ourmethod for identifying groups of transformations, for classifying semantic groups (\eg, dog breeds, plants, fruits) and recognition with the presence/absence of data type groups (\eg, \prompt{Can models recognize domestic animals with style-type transformations?}).

\subsection{Finding reproducibility analysis: data type identification}
\label{sec:bethge}

We compare the conclusions obtained by \ourmethod to the ones obtained by the manually-designed benchmark presented in~\cite{udandarao2023visual}.
Specifically, the authors created and curated two datasets (natural and synthetic) of images featuring a single animal, spanning 27 data-type transformations covering 4 broad categories: geometric (\eg, left-rotation), pixel (\eg, applying Gaussian noise), style (\eg, creating a cartoon-version), and semantic (\eg, replacing a single animal with multiple animals).
They considered both contrastively-trained VLMs, \eg, CLIP~\cite{radford2021learning}), and auto-regressively trained VLMs, \eg, LLaVA (referred to as LMMs), and asks them to identify the transformation applied on the image.
For fairness with contrastive ones, they evaluate auto-regressive VLMs using answer ranking~\cite{dai2024instructblip,lin2022truthfulqa}.
Their key finding is that VLMs excel in recognizing semantic content but fail to acquire an understanding of visual data-types.

To be comparable with \cite{udandarao2023visual}, we prompt \ourmethod with each user query per data type and let \ourmethod deploy the models, define the experiments, generate the data, test models, and analyze results.
The prompt template for formulating the $\operatorname{query}$ follows the template: \prompt{Can models identify \{data\_type\} in images?}.
For some special cases, we rephrased the query to reduce ambiguities (\eg, we replace \prompt{Can models identify multi-different in images?} with \prompt{Can models identify if multiple objects of different classes are in the image?}.
We report all the used prompts in the Appendix.

\begin{figure}
  \centering
  \includegraphics[width=1.0\linewidth, trim={0 10pt 0 0},clip]{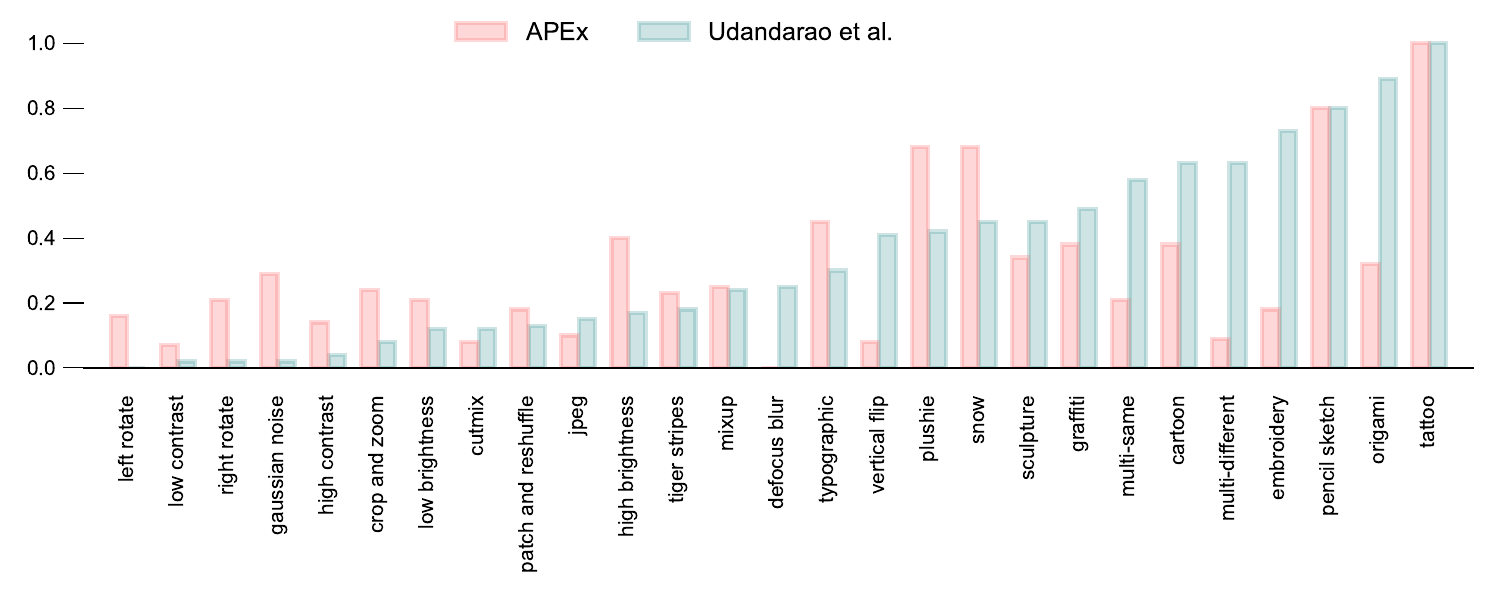}
  \caption{\textbf{Data types identification.} Summary of the normalized results (vertical axis) achieved by the models across the 27 data type identification tasks, in comparison to those obtained in~\cite{udandarao2023visual}.
  The values equal the min-max normalized performance across the set of experiments designed by \ourmethod.}
  \label{fig:data_types}
\end{figure}

The evaluation protocol of \cite{udandarao2023visual} covers all the data types together, while we test each data type independently.
As our performance metrics (average accuracy) have higher numerical values, we report the normalized metrics in Fig.~\ref{fig:data_types}, where the data types at the horizontal axis are listed in an ascending order based on the results in~\cite{udandarao2023visual}.

Despite their differences in numerical values per data type, the analysis of \ourmethod leads to a similar trend drawn from the manually designed and curated benchmark, especially when we organize the analyses by data type groups, \ie, geometric, pixel, style, and semantic~\cite{udandarao2023visual} (as shown in Tab.~\ref{tab:ranking}).
For example, both \ourmethod and \cite{udandarao2023visual} identify that LMMs perform better when recognizing {tattoos} and {embroidery} rather than lower-level transformations (\eg, {low/high contrast}, {jpeg}).
More importantly, we reach the same conclusion that LMMs are generally better in style/semantic than pixel/geometric understanding.
In the Appendix, we report the complete list of experiments \ourmethod generated for each of the 27 data types, regarding the generated questions, the answers, and the tools.

\subsection{Flexibility analyses}
\label{sec:flexibility}

\noindent\textbf{Coarse \textit{query} handling.} 
We further evaluate \ourmethod in handling $\operatorname{query}$ of a coarse granularity.
Instead of focusing on specific data type, we form the $\operatorname{query}$ using the group of transformations as in \cite{udandarao2023visual}: {geometric}, {pixel}, {semantic} and {style}.
We form general queries: \prompt{Can models identify \{transformation\_group\} transformations?} 

This study introduces multiple challenges.
First, \ourmethod has to reason on a higher level, identifying which subset of tools is relevant to answer the $\operatorname{query}$, \eg, for the {geometric} family, it should focus on geometric transformations.
Second, it must test models on multiple axes, devising experiments that regard not only the presence/absence of a data type but also the exact identification against alternatives of the same group, in a hierarchical fashion.
Similarly to the findings in Sec.~\ref{sec:bethge}, results in Tab.~\ref{tab:ranking} (expanded in Tab.~\ref{tab:data_types_group_expanded} in the Appendix) confirm the findings of \cite{udandarao2023visual} and those in Sec.~\ref{sec:bethge} where, despite different numbers in absolute terms, the trend is preserved.
The table shows that \ourmethod, just like in~\cite{udandarao2023visual}, geometric transformations (0.5 accuracy for \ourmethod, 0.1 informedness for~\cite{udandarao2023visual}) are harder to recognize w.r.t. style manipulations (0.78 accuracy for \ourmethod, 0.46 informedness for \cite{udandarao2023visual}).
Tab.~\ref{tab:data_types_groups_qual} in the Appendix, reports the quantitative analyses obtained via \ourmethod and the type of experiments conducted.
Interestingly, \ourmethod tends to experiment with multiple data types in a category (\eg, compression, noise, gray-scale for pixel-level ones).
When the objective is less clear (\eg, style-type), \ourmethod explores various comparisons with different levels of granularity, \eg, from varying both semantic and style (dog oil paintings vs bird pastel) to fixing the former and varying the latter (baroque structure vs art deco structure).

\begin{figure}[t]
    \centering
    \begin{minipage}[hb]{0.48\textwidth}
        \centering
        \vspace{15pt}
        \captionof{table}{\textbf{Data type group ranking.} Ranking of data type group recognition performance from the best (top) to worst (bottom).
        \ourmethod achieves the same ranking of~\cite{udandarao2023visual} both when testing each data type independently and aggregating metrics by their group (Avg. types), and when directly querying for the group understanding (Groups).}
    \label{tab:ranking}
    \resizebox{1.\textwidth}{!}{\begin{tabular}{c|ll|l}
    \toprule
      & \multicolumn{2}{c}{\ourmethod}& \multirow{2}{*}{~\cite{udandarao2023visual} }\\
    \textbf{Rank} & \textbf{Avg. types} & \textbf{Groups} & \\
    \midrule
    1 & Style & Style & Style \\
    2 & Semantic & Semantic & Semantic \\
    3 & Pixel & Pixel & Pixel \\
    4 & Geometric & Geometric & Geometric \\
    \bottomrule
    \end{tabular}}
    \end{minipage}
    \hfill
    \begin{minipage}[hb]{0.48\textwidth}
        \centering
\includegraphics[width=0.75\textwidth]
{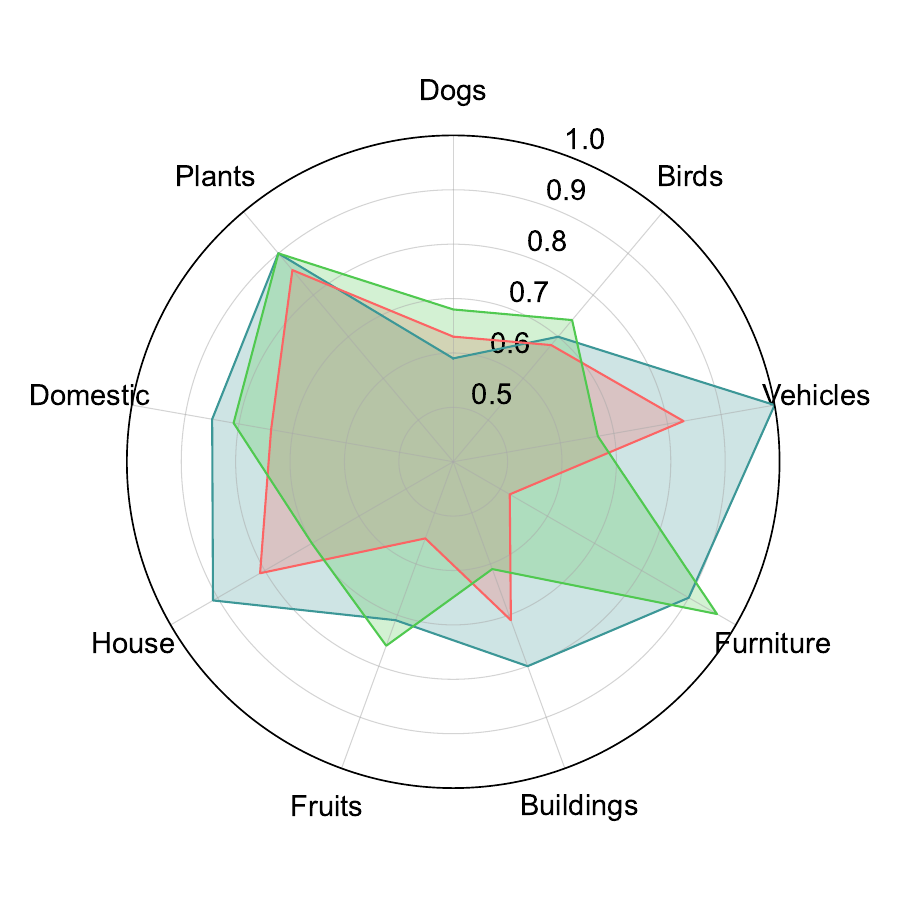}
\vspace{-15pt}
\captionof{figure}{\textbf{Data classes recognition.} Accuracy averaged over experiments of \ourmethod for \inlineColorbox{DrawioBlue}{BLIP-2}, \inlineColorbox{DrawioRed}{IDEFICS}, and \inlineColorbox{DrawioGreen}{LLaVA} on the nine recognition tasks.} 
\label{fig:data_classes_radar}
    \end{minipage}
\end{figure}

\noindent\textbf{Data classes recognition.}
To further assess the potential of \ourmethod, we challenge it on fine-grained recognition experiments, requiring precise semantic modeling.
Specifically, we ask \ourmethod to assess LMMs' performance on nine semantic groups of classes, \ie, dog breeds, bird types, vehicles, furniture, building types, fruits, household objects, domestic animals, and plants.\footnote{We select these groups from the 17 of ImageNet-X~\cite{idrissi2022imagenet}, removing too generic ones (\eg, other, commodity) and merging similar concepts (\eg, vehicle and wheeled vehicle).}
An example $\operatorname{query}$ is \prompt{Can models recognize dog breeds?}.

Fig.~\ref{fig:data_classes_radar} shows the average class-wise accuracy of the different models of each group.
The experiments show that BLIP-2 generally outperforms the other models, in particular on vehicles, buildings, and household objects.
For animals, the performance varies on the species, with LLaVA being the best on birds and dogs, while BLIP-2 outperforming it on the coarser domestic category.
We can also observe interesting phenomenon.
For instance, for birds, no model can distinguish blue jay from cardinal, and goldfinch from canary.
The only model that can distinguish cathedrals from skyscrapers is BLIP-2, achieving 0.88 accuracy.
On the other hand, it cannot separate churches from museums, achieving 0.56, with the others achieving perfect scores.
When asking about fruit colors (\ie, showing apples and bananas and asking whether they are red or yellow), LLaVA achieves random chance performance, while BLIP-2 reaches perfect score.
In addition, we report the automatically generated quantitative results, in Tab.~\ref{tab:data_classes} in the Appendix.
The table shows that \ourmethod designs interesting types of experiments, comparing classes with extremely close similarity (\eg, husky vs malamute), sometimes going from coarse comparisons to finer ones (\eg, for domestic animals, first cat vs dog, then Persian vs Siamese), and even asking for attributes (in buildings, red vs blue).

\begin{figure}
  \centering
  \includegraphics[width=1.0\linewidth,trim={0 20pt 0 0},clip]{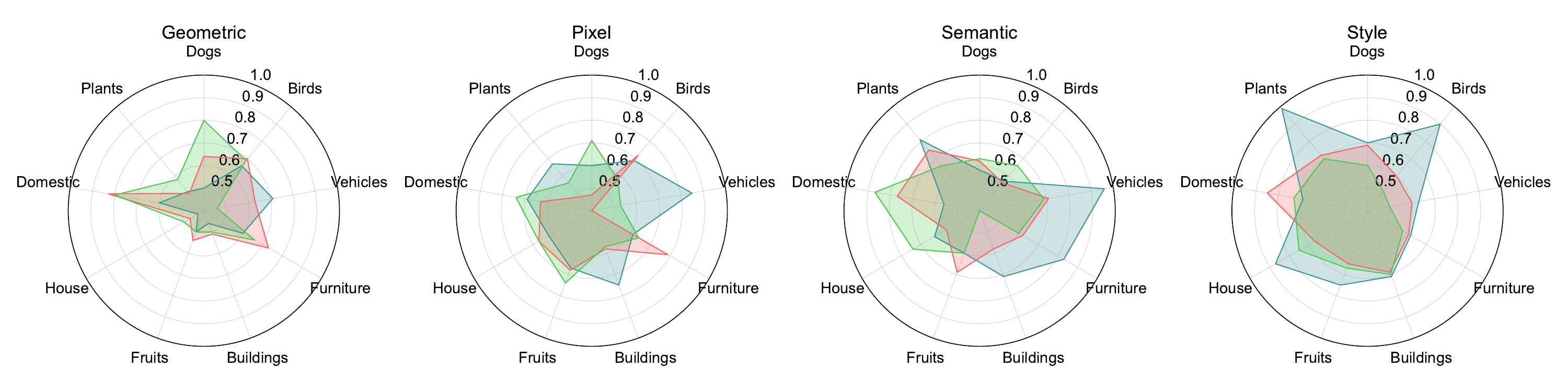}
  \caption{\textbf{Data classes robustness to data types.} Summary of the average accuracy achieved by \inlineColorbox{DrawioBlue}{BLIP-2}, \inlineColorbox{DrawioRed}{IDEFICS}, and \inlineColorbox{DrawioGreen}{LLaVA} across the nine recognition tasks when adding different data type transformations.
  Accuracy is averaged across the set of experiments designed by \ourmethod.}
  \label{fig:robustness_radar}
\end{figure}

Since \ourmethod can process arbitrary queries, even those combining multiple requests, we perform a final experiment by testing LMMs, combining semantics with data types.
Specifically, we consider the 9 semantic groups and the 4 data-types groups of the previous experiment and combine them, \ie, forming queries of the type: \prompt{Can models recognize [class] under [data type]?}.
We report the results in Fig.~\ref{fig:robustness_radar}.
Overall, we see a similar trend of Tab.~\ref{tab:ranking}, with performance improving as we move from geometric to style transformations.
Also, similarly to Fig.~\ref{fig:data_classes_radar}, BLIP-2 tends to achieve better results on objects (\eg, vehicles) while LLaVA is better with animals (\eg, dogs).
Interestingly, we see different trends based on the data types, with the best performing model often changing across transformations, \eg, fruits have IDEFICS as best for geometric- and semantic-type data, while LLaVA for pixel-type and BLIP-2 for style-type.
These patterns are easier to spot with \ourmethod, thanks to its possibility to test various configurations without manual effort.

\section{Conclusion}
\label{sec:conclusion}

The introduction of \ourmethod represents a significant advancement in LMM benchmarking.
By automating the entire benchmarking process, from design to execution and analysis, \ourmethod greatly reduces the researcher's evaluation effort, potentially reducing the risks related to subjective evaluation.
The possibility of integrating other tools and functionalities into \ourmethod makes it versatile.
Moreover, we experimentally demonstrate the reliability of \ourmethod in producing valid reports, by reproducing conclusions obtained by previous studies.
Beyond its immediate implications for LMMs, the development of \ourmethod contributes to the broader field of AI and machine learning by providing LMMs with automated, reproducible, and comprehensive benchmarking.

\textbf{Limitations.} While \ourmethod shows promising results, it inherits the limitations of visual programming methods~\cite{gupta2023visprog,suris2023vipergpt,shaham2024maia}.
Specifically, it is constrained by the reasoning capabilities of the underlying language model.
For instance, we identify failure cases due to the model inability to understand the report or the query (\eg, repetitions of questions) or due to wrong task interpretation (\eg, ambiguous answers, random transformations/selections applied).
Interesting failure cases are "meta" ones, where \ourmethod asks questions about the LMM itself rather than testing it.
Nevertheless, these issues can be mitigated by the self-healing control (see Sec.\ref{sec:implementation details}) and can be further addressed using more powerful LLMs (\eg, GPT-4), readily applicable to \ourmethod thanks to its modularity.

\bibliographystyle{apalike}
\bibliography{refs}

\newpage
\appendix

\section{Appendix}

\subsection{Additional Implementation Details}
With reference to the transformation tools described in Sec.~\ref{sec:implementation details}, the list of transforms implemented in \ourmethod are \texttt{AddGaussianNoise}, \texttt{AddJPEGCompression}, \texttt{ApplyCutMix}, \texttt{ApplyMixUp}, \texttt{ChangeBrightness}, \texttt{ChangeContrast}, \texttt{CropRandomShuffleAndRecompose}, \texttt{DefocusBlurImage}, \texttt{EditImageStyle}, \texttt{EditImageWeather}, \texttt{FlipImage}, \texttt{Identity}, \texttt{OverlayColor}, \texttt{PasteGeneratedObjectAtRandomPosition}, \texttt{PasteGeometricShapeAtRandomPosition}, \texttt{PasteTextAtRandomPosition}, \texttt{RotateImage}, and \texttt{ZoomAtRandomPosition}.
We report the full docstrings of all the tools below.
We devise the transformation tools to produce data types defined in~\cite{udandarao2023visual}.
All data types are covered, but not all are explicitly defined with a tool (\eg, we do not design specific tools for style transfer to cartoons, but provide a generic tool for style editing).

\subsection{Expanded results on data type groups ranking}

We additionally report in Tab.~\ref{tab:data_types_group_expanded} the numbers we used to rank the data type groups in Tab.~\ref{tab:ranking}.
Note that, the values reported are average accuracy for \ourmethod and informedness for~\cite{udandarao2023visual}.
For \ourmethod we consider both the average case (\ie, where we perform experiments on each data type of \cite{udandarao2023visual} and average the result by groups) and the grouped one (\ie, where we evaluate recognition directly on the grouped data types).
While the ranges are different, we can still see the same patterns across all three evaluation criteria, with the highest gaps between Style and Geometric types (\eg, 0.2 accuracy for Avg. types, 0.18 for Groups in accuracy, while 0.36 for \cite{udandarao2023visual} in informedness.
If we take two consecutive positions, for all three models, the gap between style-type and semantic-type is the highest (\ie, 0.14 for Avg., 0.18 for Groups, 0.17 for \cite{udandarao2023visual}).
The lowest gap between two consecutive positions differs instead, as Avg. and Groups show a lower gap between semantic-type and pixel-type (\ie, 0.01 and 0.02 respectively), while \cite{udandarao2023visual} shows the smallest gap between pixel and geometric transformations.
Despite this small relative difference, results are consistent, with low-level transformations (\ie, pixel, geometric) achieving lower results than higher-level ones (\ie, style, semantic).

\begin{table}
\caption{\textbf{Data type group ranking results.} 
Ranking of data type groups from easiest to hardest with related scores.
We report the numbers of the data type group averages as a complementary representation of Tab.~\ref{tab:ranking} in the main paper.}
\label{tab:data_types_group_expanded}
\centering
\begin{tabular}{l|ll|l}
\toprule
& \multicolumn{2}{c}{\ourmethod}& \multirow{2}{*}{~\cite{udandarao2023visual} }\\
\textbf{Rank} & \textbf{Avg. types} & \textbf{Groups} & \\
\midrule
1 & Style (0.74) & Style (0.68) & Style (0.46) \\
2 & Semantic (0.60) & Semantic (0.60) & Semantic (0.29) \\
3 & Pixel (0.59) & Pixel (0.58) & Pixel (0.11) \\
4 & Geometric (0.56) & Geometric (0.50) & Geometric (0.10) \\
\bottomrule
\end{tabular}
\end{table}

\subsection{Detailed results}
In this section, we provide details about the results reported in Sec.~\ref{sec:exp}.
In particular, results in Tab.~\ref{tab:data_types} are associated with results already reported in Fig.~\ref{fig:data_types}.
Similarly, Tab.~\ref{tab:data_types_groups_qual} is associated to experiments on groups of data types, discussed in the same section.
Finally, numbers reported in Table \ref{tab:data_classes} corresponds to Fig.\ref{fig:data_classes_radar}.

By analyzing the results with \ourmethod, we observe some interesting behaviors of LMMs.
For example, the recognition performance of LMMs can be extremely sensitive to the questions/answers.
Simply rephrasing the question-answers pairs, one can expect very different performances.
Examples of this behavior can be observed by looking at Tab.~\ref{tab:data_types}.
While certain sensitivity to textual prompts is expected, it is often under-explored in current studies, for instance, in the study we compared with~\cite{udandarao2023visual}, only fixed prompts are used in their benchmark.
Instead, the prompt question generation in \ourmethod is handled by an LLM that produces different prompt questions at each round of experiment design, facilitating new findings along this aspect.

There are also interesting findings specific to different LMMs.
For example (see Tab.~\ref{tab:data_types}), BLIP-2 is good at recognizing low-contrast images (up to 0.84 accuracy) while the performances of other LMMs are around random chance.
IDEFICS has a strong recognition performance on high brightness (almost perfect), while the other models underperform.
When recognizing high-contrast images, the synonym ``well-lit'' sometimes appears in the questions, but is only understood by LLaVA (with even improved performance), while other models fail to interpret the synonym properly.
Interestingly, the performance of recognizing pixel transformation does not positively correlate with their semantic understanding.
While models cannot recognize the presence of high contrast, that does not undermine their understanding of the semantic class depicted in the image.

\begin{tiny}
\setlength\LTcapwidth{\linewidth}

\begin{longtable}{llccc}
\caption{\normalsize{\textbf{Data types results.} We report the complete outputs associated with \ourmethod when queries about data types are provided as input.
Due to limited space, we refer to tools with abbreviated names.
Better seen at magnification.}} \label{tab:data_types} \\
\toprule
\rowcolor{lightgray}\multirow[t]{2}{*}{\textbf{\textsc{Q\&A}}} & \multirow[t]{2}{*}{\textbf{\textsc{Tools}}} & \multicolumn{3}{c}{\textbf{\textsc{Models}}} \\
\rowcolor{lightgray}& & \textsc{blip-2} & \textsc{idefics} & \textsc{llava} \\

\midrule
\rowcolor{beaublue}\multicolumn{5}{l}{\prompt{Can models identify left rotation in images?}} \\
\multicolumn{2}{l}{\textit{Is the image rotated to the left?}} & 0.49 & 0.47 & 0.50 \\
- Yes & Retrieval("random"), Rotate(-90) \\
- No & Retrieval("random") \\
\cmidrule(r){1-2}
\multicolumn{2}{l}{\textit{Is the wheeled vehicle in the image a car or a truck?}} & 1.00 & 0.50 & 0.77 \\
- A car & Generation("car"), Rotate(-90) \\
- A truck & Generation("truck"), Rotate(-90) \\
\cmidrule(r){1-2}
\multicolumn{2}{l}{\textit{Is the image rotated to the left?}} & 0.43 & 0.50 & 0.50 \\
- Yes & Retrieval("random"), Rotate(-90) \\
- No & Retrieval("random") \\
\cmidrule(r){1-2}
\multicolumn{2}{l}{\textit{Is the object in the image rotated to the left?}} & 0.50 & 0.66 & 0.50 \\
- Yes & Generation("dog"), Rotate(-90) \\
- No & Generation("dog") \\
\cmidrule(r){1-2}
\multicolumn{2}{l}{\textit{Does the image contain a rotated object?}} & 0.43 & 0.50 & 0.53 \\
- Yes & Retrieval("random"), Rotate(-90) \\
- No & Retrieval("random") \\

\midrule
\rowcolor{beaublue}\multicolumn{5}{l}{\prompt{Can models identify low contrast in images?}} \\
\multicolumn{2}{l}{\textit{Is the image low-contrast?}} & 0.19 & 0.48 & 0.50 \\
- Yes & Retrieval("dog"), ChangeContrast(0.5) \\
- No & Retrieval("vehicle") \\
\cmidrule(r){1-2}
\multicolumn{2}{l}{\textit{Is the contrast level of the image very low?}} & 0.84 & 0.50 & 0.50 \\
- Yes, very low contrast & Generation("dog"), ChangeContrast(0.3) \\
- No, contrast is high & Generation("vehicle"), ChangeContrast(1.0) \\

\midrule
\rowcolor{beaublue}\multicolumn{5}{l}{\prompt{Can models identify zoom in images?}} \\
\multicolumn{2}{l}{\textit{Has the image been zoomed in?}} & 0.50 & 0.46 & 0.50 \\
- Yes & Retrieval("wheeled vehicle"), Zoom(1.5) \\
- No & Retrieval("wheeled vehicle") \\
\cmidrule(r){1-2}
\multicolumn{2}{l}{\textit{Is the object in the image a bird?}} & 0.48 & 0.53 & 0.90 \\
- Yes & Retrieval("bird"), Zoom(1.5) \\
- No & Retrieval("random") \\
\cmidrule(r){1-2}
\multicolumn{2}{l}{\textit{Is the object in the image a dog?}} & 0.50 & 0.50 & 1.00 \\
- Yes & Retrieval("dog"), Zoom(1.5) \\
- No & Retrieval("bird"), Zoom(1.5) \\

\midrule
\rowcolor{beaublue}\multicolumn{5}{l}{\prompt{Can models identify right rotation in images?}} \\
\multicolumn{2}{l}{\textit{Is the image rotated to the right?}} & 0.47 & 0.55 & 0.50 \\
- Yes & Retrieval("random"), Rotate(90) \\
- No & Retrieval("random") \\
\cmidrule(r){1-2}
\multicolumn{2}{l}{\textit{Is the image rotated to the right?}} & 0.50 & 0.85 & 0.50 \\
- Yes & Retrieval("wheeled vehicle"), Rotate(90) \\
- No & Retrieval("device") \\
\cmidrule(r){1-2}
\multicolumn{2}{l}{\textit{Is the rotated image showing a wheeled vehicle?}} & 0.61 & 0.50 & 1.00 \\
- Yes & Retrieval("wheeled vehicle"), Rotate(90) \\
- No & Retrieval("device") \\
\cmidrule(r){1-2}
\multicolumn{2}{l}{\textit{Is the image rotated to the right?}} & 0.41 & 0.67 & 0.50 \\
- Yes & Retrieval("device"), Rotate(180) \\
- No & Retrieval("device") \\
\cmidrule(r){1-2}
\multicolumn{2}{l}{\textit{Is the image rotated to the right by 45 degrees?}} & 0.60 & 0.50 & 0.50 \\
- Yes & Retrieval("device"), Rotate(45) \\
- No & Retrieval("device") \\

\midrule
\rowcolor{beaublue}\multicolumn{5}{l}{\prompt{Can models identify Gaussian noise in images?}} \\
\multicolumn{2}{l}{\textit{Is there gaussian noise in the image?}} & 0.65 & 0.50 & 1.00 \\
- Yes & Generation("random"), AddGaussianNoise(1.4) \\
- No & Generation("random") \\
\cmidrule(r){1-2}
\multicolumn{2}{l}{\textit{Does the image contain significant gaussian noise?}} & 0.51 & 0.50 & 0.97 \\
- Yes & Generation("random"), AddGaussianNoise(2.0) \\
- No & Generation("random") \\
\cmidrule(r){1-2}
\multicolumn{2}{l}{\textit{Does the image contain mild gaussian noise?}} & 0.70 & 0.50 & 0.79 \\
- Yes & Generation("photography"), AddGaussianNoise(1.7) \\
- No & Generation("photography") \\
\cmidrule(r){1-2}
\multicolumn{2}{l}{\textit{Is the image corrupted with mild gaussian noise?}} & 0.57 & 0.56 & 0.52 \\
- Yes & Generation("random"), AddGaussianNoise(1.7) \\
- No & Generation("random") \\
\cmidrule(r){1-2}
\multicolumn{2}{l}{\textit{Is the image corrupted with mild gaussian noise?}} & 0.51 & 0.53 & 0.53 \\
- Yes & Generation("random"), AddGaussianNoise(1.7) \\
- No & Generation("random") \\

\midrule
\rowcolor{beaublue}\multicolumn{5}{l}{\prompt{Can models identify high contrast in images?}} \\
\multicolumn{2}{l}{\textit{Is the image high contrast?}} & 0.49 & 0.48 & 0.50 \\
- Yes & Generation("random"), ChangeContrast(2) \\
- No & Generation("random"), ChangeContrast(2) \\
\cmidrule(r){1-2}
\multicolumn{2}{l}{\textit{Is the image a high contrast scene?}} & 0.61 & 0.50 & 0.76 \\
- Yes & Generation("high contrast scene"), ChangeContrast(2) \\
- No & Generation("low contrast scene"), ChangeContrast(0.5) \\
\cmidrule(r){1-2}
\multicolumn{2}{l}{\textit{Is the image well-lit?}} & 0.75 & 0.50 & 0.50 \\
- Yes & Retrieval("structure"), ChangeContrast(2) \\
- No & Retrieval("natural object"), ChangeContrast(0.5) \\
\cmidrule(r){1-2}
\multicolumn{2}{l}{\textit{Is the object in the image clearly visible in a high contrast scene?}} & 0.51 & 0.50 & 0.50 \\
- Yes & Generation("structure"), ChangeContrast(2) \\
- No & Generation("natural object"), ChangeContrast(0.5) \\
\cmidrule(r){1-2}
\multicolumn{2}{l}{\textit{Is the vehicle in the image clearly visible in a high contrast scene?}} & 0.53 & 0.50 & 0.50 \\
- Yes & Generation("wheeled vehicle"), ChangeContrast(2) \\
- No & Generation("wheeled vehicle"), ChangeContrast(0.5) \\

\midrule
\rowcolor{beaublue}\multicolumn{5}{l}{\prompt{Can models identify low brightness in images?}} \\
\multicolumn{2}{l}{\textit{Is the image in low brightness?}} & 0.39 & 0.51 & 0.50 \\
- Yes & Generation("random"), ChangeBrightness(0.5) \\
- No & Generation("random") \\
\cmidrule(r){1-2}
\multicolumn{2}{l}{\textit{Is the image in very low brightness?}} & 0.33 & 0.50 & 0.50 \\
- Yes & Generation("random"), ChangeBrightness(0.2) \\
- No & Generation("random") \\
\cmidrule(r){1-2}
\multicolumn{2}{l}{\textit{Is the image in extremely low brightness?}} & 0.52 & 0.50 & 0.50 \\
- Yes & Generation("random"), ChangeBrightness(0.1) \\
- No & Generation("random") \\
\cmidrule(r){1-2}
\multicolumn{2}{l}{\textit{Is the image in extremely low brightness?}} & 0.46 & 0.50 & 0.50 \\
- Yes & Generation("random"), ChangeBrightness(0.1) \\
- No & Generation("random") \\
\cmidrule(r){1-2}
\multicolumn{2}{l}{\textit{What object is present in the image in extremely low brightness?}} & 1.00 & 1.00 & 1.00 \\
- A cat & Generation("cat"), ChangeBrightness(0.1) \\
- A car & Generation("car"), ChangeBrightness(0.1) \\

\midrule
\rowcolor{beaublue}\multicolumn{5}{l}{\prompt{Can models identify the presence of patching and reshuffling in images?}}  \\
\multicolumn{2}{l}{\textit{Is the image randomly reshuffled?}} & 0.42 & 0.50 & 0.83 \\
- No & Retrieval("random") \\
- Yes & Retrieval("random"), CropAndShuffle(2) \\
\cmidrule(r){1-2}
\multicolumn{2}{l}{\textit{Is the image randomly patched?}} & 0.53 & 0.42 & 0.79 \\
- No & Retrieval("random") \\
- Yes & Retrieval("random"), \\ & PasteGeometricShape("square", 48, [255, 0, 0], False, 1) \\
\cmidrule(r){1-2}
\multicolumn{2}{l}{\textit{Does the image contain a randomly patched object or reshuffled patches?}} & 0.54 & 0.52 & 0.50 \\
- No & Retrieval("random") \\
- Yes & Retrieval("random"), CropAndShuffle(2) \\
\cmidrule(r){1-2}
\multicolumn{2}{l}{\textit{Does the image contain a randomly patched object and randomly reshuffled patches?}} & 0.60 & 0.50 & 0.82 \\
- No & Retrieval("random") \\
- Yes & Retrieval("random"), CropAndShuffle(3) \\
\cmidrule(r){1-2}
\multicolumn{2}{l}{\textit{Is the image non-randomly reshuffled and patched?}} & 0.67 & 0.32 & 0.50 \\
- No & Retrieval("random") \\
- Yes & Retrieval("random"), Mixup(1.0) \\

\midrule
\rowcolor{beaublue}\multicolumn{5}{l}{\prompt{Can models identify the presence of cutmix in images?} } \\
\multicolumn{2}{l}{\textit{Is cutmix present in the image?}} & 0.61 & 0.51 & 0.62 \\
- Yes & Retrieval("random"), CutMix(alpha=1.0) \\
- No & Retrieval("random") \\
\cmidrule(r){1-2}
\multicolumn{2}{l}{\textit{What is the size of the cutmix present in the image?}} & 0.35 & 0.30 & 0.33 \\
- Small & Retrieval("cutmix"), CutMix(0.5) \\
- Large & Retrieval("cutmix"), CutMix(1.5) \\
- None & Retrieval("random") \\
\cmidrule(r){1-2}
\multicolumn{2}{l}{\textit{Is the cutmix covering a large portion of the image?}} & 0.42 & 0.50 & 0.92 \\
- Yes & Generation("cutmix") \\
- No & Retrieval("random") \\

\midrule
\rowcolor{beaublue}\multicolumn{5}{l}{\prompt{Can models identify jpeg compression in images?}} \\
\multicolumn{2}{l}{\textit{Is the image compressed with jpeg compression?}} & 0.50 & 0.31 & 0.50 \\
- Yes & Retrieval("wheeled vehicle"), AddJPEGCompression(18.0) \\
- No & Retrieval("dog") \\
\cmidrule(r){1-2}
\multicolumn{2}{l}{\textit{Is the image heavily compressed with jpeg compression?}} & 0.50 & 0.50 & 0.50 \\
- Yes & Generation("wheeled vehicle"), AddJPEGCompression(15.0) \\
- No & Retrieval("dog") \\
\cmidrule(r){1-2}
\multicolumn{2}{l}{\textit{Is the image moderately compressed with jpeg compression?}} & 0.50 & 0.50 & 0.50 \\
- Yes & Generation("dog"), AddJPEGCompression(20.0) \\
- No & Retrieval("wheeled vehicle") \\
\cmidrule(r){1-2}
\multicolumn{2}{l}{\textit{Is the image lightly compressed with jpeg compression?}} & 0.92 & 0.50 & 0.50 \\
- Yes & Retrieval("wheeled vehicle"), AddJPEGCompression(30.0) \\
- No & Retrieval("dog") \\

\midrule
\rowcolor{beaublue}\multicolumn{5}{l}{\prompt{Can models identify the presence of mixup in images?} } \\
\multicolumn{2}{l}{\textit{Does the image contain mixup?}} & 0.63 & 0.55 & 0.50 \\
- Yes & Retrieval("wheeled vehicle"), Mixup(0.5) \\
- No & Retrieval("wheeled vehicle") \\
\cmidrule(r){1-2}
\multicolumn{2}{l}{\textit{Does the image contain mixup?}} & 0.69 & 0.55 & 0.68 \\
- Yes & Retrieval("random"), Mixup(0.7) \\
- No & Retrieval("random") \\

\midrule
\rowcolor{beaublue}\multicolumn{5}{l}{\prompt{Can models identify high brightness in images?}} \\
\multicolumn{2}{l}{\textit{Is the brightness level in the image high?}} & 0.53 & 1.00 & 0.50 \\
- Yes & Retrieval("sunny outdoor scenes"), ChangeBrightness(2) \\
- No & Retrieval("low-light indoor scene"), ChangeBrightness(0.5) \\
\cmidrule(r){1-2}
\multicolumn{2}{l}{\textit{Is the brightness level in the image extremely high?}} & 0.62 & 1.00 & 0.50 \\
- Yes & Retrieval("sunny outdoor scenes"), ChangeBrightness(4) \\
- No & Retrieval("low-light indoor scene"), ChangeBrightness(0.5) \\
\cmidrule(r){1-2}
\multicolumn{2}{l}{\textit{Is the image extremely bright?}} & 0.50 & 0.97 & 0.50 \\
- Yes & Generation("sunny outdoor scenes"), ChangeBrightness(4) \\
- No & Generation("low-light indoor scene"), ChangeBrightness(0.5) \\

\midrule
\rowcolor{beaublue}\multicolumn{5}{l}{\prompt{Can models identify if text is present in images?} } \\
\multicolumn{2}{l}{\textit{Is there visible text in the image?}} & 0.50 & 0.77 & 0.69 \\
- Yes & Retrieval("random") \\
- No & Retrieval("random"), OverlayColor([0, 0, 0], 1.0) \\
\cmidrule(r){1-2}
\multicolumn{2}{l}{\textit{Is the word 'stop' present in the image?}} & 0.81 & 0.78 & 0.73 \\
- Yes & Generation("Stop") \\
- No & Generation("random") \\
\cmidrule(r){1-2}
\multicolumn{2}{l}{\textit{Is the word 'danger' present in the image?}} & 0.66 & 0.96 & 1.00 \\
- Yes & Generation("danger") \\
- No & Generation("random") \\
\cmidrule(r){1-2}
\multicolumn{2}{l}{\textit{Is the word 'caution' present in the image?}} & 0.48 & 0.78 & 0.91 \\
- Yes & Generation("caution") \\
- No & Generation("random") \\
\cmidrule(r){1-2}
\multicolumn{2}{l}{\textit{Is there any visible text present in the image?}} & 0.50 & 0.50 & 0.54 \\
- Yes & Retrieval("random") \\
- No & Retrieval("random"), OverlayColor([0, 0, 0], 1.0) \\

\midrule
\rowcolor{beaublue}\multicolumn{5}{l}{\prompt{Can models identify if objects are styled like plushies?} } \\
\multicolumn{2}{l}{\textit{Is the image styled like a plushie?}} & 1.00 & 0.50 & 0.50 \\
- Yes & Generation("plushie") \\
- No & Generation("dog") \\
\cmidrule(r){1-2}
\multicolumn{2}{l}{\textit{Is the object in the image styled like a plushie dog or a plushie cat?}} & 1.00 & 1.00 & 1.00 \\
- A plushie dog & Generation("plushie dog") \\
- A plushie cat & Generation("plushie cat") \\
\cmidrule(r){1-2}
\multicolumn{2}{l}{\textit{Is the image styled like a plushie dog or a plushie cat?}} & 1.00 & 1.00 & 1.00 \\
- A plushie dog & Generation("plushie dog") \\
- A plushie cat & Generation("plushie cat") \\
\cmidrule(r){1-2}
\multicolumn{2}{l}{\textit{Is the object in the image styled like a plushie dog, a plushie cat, or a plushie bear?}} & 1.00 & 0.93 & 1.00 \\
- A plushie dog & Generation("plushie dog") \\
- A plushie cat & Generation("plushie cat") \\
- A plushie bear & Generation("plushie bear") \\
\cmidrule(r){1-2}
\multicolumn{2}{l}{\textit{Is the image styled like a floral plushie or a plushie dog?}} & 0.50 & 0.50 & 0.50 \\
- A floral plushie & Generation("floral plushie") \\
- A plushie dog & Generation("plushie dog") \\

\midrule
\rowcolor{beaublue}\multicolumn{5}{l}{\prompt{Can models identify tiger stripes in images?}} \\
\multicolumn{2}{l}{\textit{Are the stripes on the animal in the image tiger stripes?}} & 0.23 & 0.50 & 0.50 \\
- Yes & Retrieval("tiger") \\
- No & Retrieval("zebra") \\
\cmidrule(r){1-2}
\multicolumn{2}{l}{\textit{Do the stripes on the animal in the image resemble tiger stripes?}} & 1.00 & 0.75 & 0.52 \\
- Yes & Retrieval("tiger") \\
- No & Retrieval("zebra") \\
\cmidrule(r){1-2}
\multicolumn{2}{l}{\textit{Is the image displaying tiger stripes?}} & 0.50 & 0.73 & 0.54 \\
- Yes & Retrieval("tiger") \\
- No & Retrieval("zebra") \\

\midrule
\rowcolor{beaublue}\multicolumn{5}{l}{\prompt{Can models identify snow in images?}} \\
\multicolumn{2}{l}{\textit{Is there snow in the image?}} & 0.50 & 0.50 & 1.00 \\
- Yes & Retrieval("snowy landscape") \\
- No & Retrieval("landscape") \\
\cmidrule(r){1-2}
\multicolumn{2}{l}{\textit{What is the dominant color of the snowy region in the image?}} & 1.00  & 1.00 & 0.96 \\
- White & Retrieval("snowy landscape") \\
- Blue & Retrieval("snowy landscape"), OverlayColor([0, 0, 255], 1.0) \\

\midrule
\rowcolor{beaublue}\multicolumn{5}{l}{\prompt{Can models identify graffiti-styled images?}} \\
\multicolumn{2}{l}{\textit{Is the image graffiti-styled?}} & 0.54 & 0.50 & 1.00 \\
- Yes & Generation("graffiti") \\
- No & Retrieval("random") \\
\cmidrule(r){1-2}
\multicolumn{2}{l}{\textit{What type of graffiti style is depicted in the image?}} & 0.96 & 0.50 & 0.50 \\
- Calligraffiti & Retrieval("calligraffiti") \\
- Stencil graffiti & Retrieval("stencil graffiti") \\

\midrule
\rowcolor{beaublue}\multicolumn{5}{l}{\prompt{Can models identify vertical flip in images?}} \\
\multicolumn{2}{l}{\textit{Is the image flipped vertically?}} & 0.50 & 0.50 & 0.50 \\
- Yes & Generation("wheeled vehicle"), Flip("vertical") \\
- No & Generation("wheeled vehicle") \\
\cmidrule(r){1-2}
\multicolumn{2}{l}{\textit{Is there a vertical flip in the image of the wheeled vehicle?}} & 0.50 & 0.50 & 0.50 \\
- Yes & Generation("wheeled vehicle"), Flip("vertical") \\
- No & Generation("wheeled vehicle") \\
\cmidrule(r){1-2}
\multicolumn{2}{l}{\textit{Is the vehicle in the image flipped vertically?}} & 0.50 & 0.50 & 0.50 \\
- Yes & Generation("wheeled vehicle"), Flip("vertical") \\
- No & Generation("wheeled vehicle") \\
\cmidrule(r){1-2}
\multicolumn{2}{l}{\textit{Is there a vertical flip in the image of the selected class?}} & 0.50 & 0.59 & 0.50 \\
- Yes & Generation("wheeled vehicle"), Flip("vertical") \\
- No & Generation("wheeled vehicle") \\
\cmidrule(r){1-2}
\multicolumn{2}{l}{\textit{Is the vehicle in the image flipped vertically?}} & 0.49 & 0.50 & 0.50 \\
- Yes & Generation("car"), Flip("vertical") \\
- No & Generation("car") \\

\midrule
\rowcolor{beaublue}\multicolumn{5}{l}{\prompt{Can models identify defocus blur in images?}} \\
\multicolumn{2}{l}{\textit{Is the image defocused?}} & 0.46 & 0.50 & 0.50 \\
- Yes & Generation("car"), DefocusBlur(5.0) \\
- No & Generation("car") \\
\cmidrule(r){1-2}
\multicolumn{2}{l}{\textit{Does the image contain strong defocus blur?}} & 0.37 & 0.50 & 0.50 \\
- Yes & Generation("wheeled vehicle"), DefocusBlur(8.0) \\
- No & Generation("device") \\
\cmidrule(r){1-2}
\multicolumn{2}{l}{\textit{'Is the image heavily defocused?}} & 0.38 & 0.50 & 0.50 \\
- Yes & Generation("wheeled vehicle"), DefocusBlur(8.0) \\
- No & Generation("device") \\

\midrule
\rowcolor{beaublue}\multicolumn{5}{l}{\prompt{Can models identify cartoon-styled images?}} \\
\multicolumn{2}{l}{\textit{Is the image a cartoon or a real-life photo?}} & 0.50 & 0.50 & 0.50 \\
- A cartoon & Generation("bird", "cartoon") \\
- A real-life photo & Retrieval("random") \\
\cmidrule(r){1-2}
\multicolumn{2}{l}{\textit{What type of cartoon is the image depicting?}} & 0.50 & 0.50 & 0.50 \\
- A cartoon of a bird & Generation("bird", "cartoon") \\
- A real-life photo of a bird & Retrieval("bird") \\
\cmidrule(r){1-2}
\multicolumn{2}{l}{\textit{Is the image depicting a comic-style illustration of a bird or a dog?}} & 1.00 & 1.00 & 1.00 \\
- A bird & Generation("bird", "comic-style") \\
- A dog & Generation("dog", "comic-style") \\

\midrule
\rowcolor{beaublue}\multicolumn{5}{l}{\prompt{Can models identify if several identical objects are present in images?}} \\
\multicolumn{2}{l}{\textit{Is the image showing a single identical object?}} & 0.50 & 0.71 & 0.87 \\
- Yes & Generation("car") \\
- No & Retrieval("random") \\
\cmidrule(r){1-2}
\multicolumn{2}{l}{\textit{Are there multiple identical objects in the image?}} & 0.50 & 0.50 & 0.50 \\
- Yes & Generation("car") \\
- No & Retrieval("random") \\
\cmidrule(r){1-2}
\multicolumn{2}{l}{\textit{Are there multiple identical objects of different sizes and orientations in the image?}} & 0.50 & 0.50 & 0.50 \\
- Yes & Generation("car"), PasteObject("car", 64, 3) \\
- No & Retrieval("random") \\
\cmidrule(r){1-2}
\multicolumn{2}{l}{\textit{How many identical objects are present in the image?}} & 0.50 & 0.50 & 0.72 \\
- One & Retrieval("random") \\
- Two & Generation("car"), PasteObject("car", 64, 2) \\
\cmidrule(r){1-2}
\multicolumn{2}{l}{\textit{How many identical objects are present in the image?}} & 0.50 & 0.50 & 0.91 \\
- One & Retrieval("car") \\
- Two & Retrieval("random"), PasteObject("car", 64, 2) \\

\midrule
\rowcolor{beaublue}\multicolumn{5}{l}{\prompt{Can models identify embroidery-styled images?}} \\
\multicolumn{2}{l}{\textit{Is the image in the embroidery style?}} & 0.56 & 0.50 & 0.91 \\
- Yes & Generation("embroidery") \\
- No & Retrieval("random") \\
\cmidrule(r){1-2}
\multicolumn{2}{l}{\textit{What object is depicted in the embroidery-styled image?}} & 0.50 & 0.50 & 0.50 \\
- A bonsai tree & Generation("bonsai tree", "embroidery") \\
- Another object & Retrieval("random", "embroidery") \\
\cmidrule(r){1-2}
\multicolumn{2}{l}{\textit{Is the object in the embroidery-styled image a bonsai tree or a flower?}} & 0.50 & 0.50 & 0.50 \\
- A bonsai tree & Generation("bonsai tree", "embroidery") \\
- A flower & Generation("flower", "embroidery") \\
\cmidrule(r){1-2}
\multicolumn{2}{l}{\textit{Is the object in the embroidery-styled image a flower or a butterfly?}} & 1.00 & 0.50 & 0.50 \\
- A flower & Generation("flower", "embroidery") \\
- A butterfly & Generation("butterfly", "embroidery") \\
\cmidrule(r){1-2}
\multicolumn{2}{l}{\textit{What object is depicted in the embroidery-styled image?}} & 0.50 & 0.50 & 0.50 \\
- A bonsai tree & Generation("bonsai tree", "embroidery") \\
- Another object & Retrieval("random", "embroidery") \\

\midrule
\rowcolor{beaublue}\multicolumn{5}{l}{\prompt{Can models identify if multiple objects of different classes are in the image?} } \\
\multicolumn{2}{l}{\textit{Is there more than one dog in the image?}} & 0.50 & 0.42 & 0.73 \\
- Yes & Generation("dog"), PasteObject("dog", 256, 2) \\
- No & Generation("dog") \\
\cmidrule(r){1-2}
\multicolumn{2}{l}{\textit{Is there more than one dog in the image?}} & 0.50 & 0.38 & 0.67 \\
- Yes & Generation("dog"), PasteObject("dog", 256, 2) \\
- No & Generation("dog") \\
\cmidrule(r){1-2}
\multicolumn{2}{l}{\textit{Are there vehicles and animals in the image?}} & 0.50 & 0.50 & 0.50 \\
- Yes & Generation("wheeled vehicle"), PasteObject("dog", 256, 1) \\
- No & Generation("device"), PasteObject("dog", 256, 1) \\
\cmidrule(r){1-2}
\multicolumn{2}{l}{\textit{Is there a car and a truck in the image? if so, are they both present?}} & 0.50 & 0.50 & 0.50 \\
- Yes & Generation("car") \\
- No & Generation("truck") \\

\midrule
\rowcolor{beaublue}\multicolumn{5}{l}{\prompt{Can models identify origami-styled in images?}} \\
\multicolumn{2}{l}{\textit{Is the image origami-styled?}} & 0.49 & 0.56 & 0.97 \\
- Yes & Generation("origami") \\
- No & Retrieval("random") \\
\cmidrule(r){1-2}
\multicolumn{2}{l}{\textit{What type of origami object is in the image?}} & 0.81 & 0.52 & 0.50 \\
- Paper crane & Generation("paper crane") \\
- Random origami & Retrieval("origami") \\
\cmidrule(r){1-2}
\multicolumn{2}{l}{\textit{What type of origami object is in the image?}} & 0.82 & 0.50 & 0.50 \\
- Paper crane & Generation("paper crane") \\
- Random origami & Retrieval("origami") \\
\cmidrule(r){1-2}
\multicolumn{2}{l}{\textit{Is the image a paper crane?}} & 0.45 & 0.50 & 0.50 \\
- Yes & Generation("paper crane") \\
- No & Retrieval("origami") \\
\cmidrule(r){1-2}
\multicolumn{2}{l}{\textit{Is the image showing a folded paper boat?}} & 0.61 & 0.50 & 0.50 \\
- Yes & Generation("paper boat") \\
- No & Generation("random") \\

\midrule
\rowcolor{beaublue}\multicolumn{5}{l}{\prompt{Can models identify sketch-styled in images?}} \\
\multicolumn{2}{l}{\textit{Is the image a pencil sketch of a vehicle or a natural object?}} & 0.52 & 1.00 & 1.00 \\
- Yes & Retrieval("wheeled vehicle"), EditStyle("pencil sketch") \\
- No & Retrieval("snake"), EditStyle("pencil sketch") \\
\cmidrule(r){1-2}
\multicolumn{2}{l}{\textit{Is the image a pencil sketch of a bird or a primate?}} & 0.94 & 0.91 & 0.97 \\
- A bird & Retrieval("bird"), EditStyle("pencil sketch") \\
- A primate & Retrieval("primate"), EditStyle("pencil sketch") \\

\midrule
\rowcolor{beaublue}\multicolumn{5}{l}{\prompt{Can models identify sculpture-styled in images?}} \\
\multicolumn{2}{l}{\textit{Is the image sculpture-styled?}} & 0.41 & 0.50 & 1.00 \\
- Yes & Retrieval("sculpture"), EditStyle("sculpture") \\
- No & Retrieval("random") \\
\cmidrule(r){1-2}
\multicolumn{2}{l}{\textit{Is the sculpture in the image a stone sculpture?}} & 0.76 & 0.57 & 0.98 \\
- Yes & Generation("sculpture"), EditStyle("stone") \\
- No & Retrieval("random") \\
\cmidrule(r){1-2}
\multicolumn{2}{l}{\textit{Is the sculpture in the image a stone sculpture?}} & 0.68 & 0.54 & 0.97 \\
- Yes & Generation("sculpture"), EditStyle("stone") \\
- No & Retrieval("random") \\
\cmidrule(r){1-2}
\multicolumn{2}{l}{\textit{Is the sculpture in the image a marble sculpture?}} & 0.72 & 0.56 & 0.97 \\
- Yes & Generation("marble sculpture") \\
- No & Retrieval("random") \\
\cmidrule(r){1-2}
\multicolumn{2}{l}{\textit{Is the sculpture in the image a bronze sculpture?}} & 0.48 & 0.15 & 0.50 \\
- Yes & Generation("bronze") \\
- No & Retrieval("sculpture"), EditStyle("bronze") \\

\midrule
\rowcolor{beaublue}\multicolumn{5}{l}{\prompt{Can models identify tattoo-styled in images?} } \\
\multicolumn{2}{l}{\textit{What kind of tattoo is depicted in the image?}} & 1.00 & 1.00 & 1.00 \\
- A tribal tattoo & Retrieval("tribal tattoo"), EditStyle("tattoos") \\
- A watercolor tattoo & Retrieval("watercolor tattoo"), EditStyle("tattoos") \\
\cmidrule(r){1-2}
\multicolumn{2}{l}{\textit{What kind of tattoo style is depicted in the image?}} & 1.00 & 1.00 & 1.00 \\
- A watercolor tattoo & Retrieval("watercolor tattoo"), EditStyle("tattoos") \\
- A tribal tattoo & Retrieval("tribal tattoo"), EditStyle("tattoos") \\

\bottomrule
\end{longtable}

\end{tiny}

\begin{tiny}
\setlength\LTcapwidth{\linewidth}

\begin{longtable}{llrrr}
\caption{\normalsize{\textbf{Data types groups results.} We report the complete outputs associated with \ourmethod when considering queries about data type groups.
Due to limited space, we refer to tools with abbreviated names.
Better seen at magnification.}} \label{tab:data_types_groups_qual} \\
\toprule
\rowcolor{lightgray}\multirow[t]{2}{*}{\textbf{\textsc{Q\&A}}} & \multirow[t]{2}{*}{\textbf{\textsc{Tools}}} & \multicolumn{3}{c}{\textbf{\textsc{Models}}} \\
\rowcolor{lightgray}& & \textsc{blip-2} & \textsc{idefics} & \textsc{llava} \\

\midrule
\rowcolor{beaublue}\multicolumn{5}{l}{\prompt{Can models identify geometric-type transformations?}} \\
\multicolumn{2}{l}{\textit{Is the image rotated by 90 degrees clockwise?}} & 0.47 & 0.45 & 0.50 \\
- Yes & Retrieval("random"), Rotate(90) \\
- No & Retrieval("random") \\
\cmidrule(r){1-2}
\multicolumn{2}{l}{\textit{Is the image reflected across the vertical axis?}} & 0.52 & 0.52 & 0.50 \\
- Yes & Retrieval("random"), Flip("vertical") \\
- No & Retrieval("random") \\
\cmidrule(r){1-2}
\multicolumn{2}{l}{\textit{Is the image containing visible text?}} & 0.57 & 0.12 & 0.85 \\
- Yes & Retrieval("text") \\
- No & Retrieval("random") \\
\cmidrule(r){1-2}
\multicolumn{2}{l}{\textit{Is the image noise-corrupted?}} & 0.51 & 0.54 & 0.52 \\
- Yes & Retrieval("random"), AddGaussianNoise(1.4) \\
- No & Retrieval("random") \\
\cmidrule(r){1-2}
\multicolumn{2}{l}{\textit{Is there a zoomed-in region in the image?}} & 0.42 & 0.50 & 0.50 \\
- Yes & Retrieval("random"), Zoom(2.0) \\
- No & Retrieval("random") \\

\midrule
\rowcolor{beaublue}\multicolumn{5}{l}{\prompt{Can models identify pixel-type transformations?}} \\
\multicolumn{2}{l}{\textit{Is the image grayscale?}} & 0.93 & 0.50 & 0.50 \\
- No & Generation("dog") \\
- Yes & Generation("dog"), EditStyle("grayscale") \\
\cmidrule(r){1-2}
\multicolumn{2}{l}{\textit{Is the image rotated clockwise?}} & 0.49 & 0.48 & 0.50 \\
- No & Retrieval("random") \\
- Yes & Retrieval("random"), Rotate(90) \\
\cmidrule(r){1-2}
\multicolumn{2}{l}{\textit{Is the image affected by gaussian noise?}} & 0.56 & 0.50 & 0.66 \\
- No & Retrieval("random") \\
- Yes & Retrieval("random"), AddGaussianNoise(2) \\
\cmidrule(r){1-2}
\multicolumn{2}{l}{\textit{Is the image affected by jpeg compression?}} & 0.54 & 0.50 & 0.50 \\
- No & Retrieval("random") \\
- Yes & Retrieval("random"), AddJPEGCompression(20.0) \\
\cmidrule(r){1-2}
\multicolumn{2}{l}{\textit{Is the image affected by defocus blur?}} & 0.48 & 0.50 & 0.50 \\
- No & Retrieval("random") \\
- Yes & DefocusBlurImage("random"), AddJPEGCompression(10.0) \\

\midrule
\rowcolor{beaublue}\multicolumn{5}{l}{\prompt{Can models identify semantic-type transformations?}} \\
\multicolumn{2}{l}{\textit{Is the object in the image a bird or a vehicle?}} & 0.75 & 0.50 & 0.50 \\
- A bird & Generation("bird") \\
- A vehicle & Generation("wheeled vehicle") \\
\cmidrule(r){1-2}
\multicolumn{2}{l}{\textit{Is the background of the image a natural or man-made environment?}} & 0.50 & 0.50 & 0.50 \\
- Natural & Generation("forest") \\
- Man-made & Generation("cityscape") \\
\cmidrule(r){1-2}
\multicolumn{2}{l}{\textit{Is the style of the object in the image a photo or a pencil sketch?}} & 0.50 & 0.50 & 0.50 \\
- Photo & Generation("random", "photo") \\
- Pencil sketch & Generation("random", "pencil sketch") \\
\cmidrule(r){1-2}
\multicolumn{2}{l}{\textit{Is the weather in the image sunny or cloudy?}} & 0.74 & 0.77 & 0.97 \\
- Sunny & Retrieval("random"), EditWeather("sunny") \\
- Cloudy & Retrieval("random"), EditWeather("cloudy") \\
\cmidrule(r){1-2}
\multicolumn{2}{l}{\textit{Is the image a photograph of a vehicle or an illustration of an animal?}} & 0.80 & 0.50 & 0.50 \\
- Vehicle & Generation("wheeled vehicle") \\
- Animal illustration & Generation("random", "pencil sketch") \\

\midrule
\rowcolor{beaublue}\multicolumn{5}{l}{\prompt{Can models identify style-type transformations?}} \\
\multicolumn{2}{l}{\textit{Is the image in a specific artistic style?}} & 0.50 & 0.56 & 0.50 \\
- Painting & Generation("dog", "oil painting") \\
- Pastel & Retrieval("bird", "pastel") \\
\cmidrule(r){1-2}
\multicolumn{2}{l}{\textit{Is the image in a specific artistic style?}} & 1.00 & 1.00 & 0.50 \\
- Sketch & Generation("dog", "sketch") \\
- Mosaic & Generation("bird", "mosaic") \\
\cmidrule(r){1-2}
\multicolumn{2}{l}{\textit{Is the image in a specific artistic style?}} & 0.50 & 0.59 & 0.50 \\
- Painting & Generation("dog", "oil painting") \\
- Pastel & Retrieval("bird", "pastel") \\
\cmidrule(r){1-2}
\multicolumn{2}{l}{\textit{Does the image contain a specific geometrical shape?}} & 1.00 & 1.00 & 1.00 \\
- A circle & Generation("circle") \\
- A triangle & Generation("triangle") \\
\cmidrule(r){1-2}
\multicolumn{2}{l}{\textit{Is the image in a specific architectural style?}} & 1.00 & 1.00 & 1.00 \\
- Baroque & Retrieval("structure"), EditStyle("baroque") \\
- Art deco & Retrieval("structure"), EditStyle("art deco") \\

\bottomrule
\end{longtable}

\end{tiny}

\begin{tiny}
\begin{longtable}{llrrr}
\caption{\normalsize{\textbf{Data classes results.} We report the complete outputs associated with \ourmethod when queries about data classes are provided as input.
Due to limited space, we refer to tools with abbreviated names.
Better seen at magnification.}} \label{tab:data_classes} \\
\toprule
\rowcolor{lightgray}\multirow[t]{2}{*}{\textbf{\textsc{Q\&A}}} & \multirow[t]{2}{*}{\textbf{\textsc{Tools}}} & \multicolumn{3}{c}{\textbf{\textsc{Models}}} \\
\rowcolor{lightgray}& & \textsc{blip-2} & \textsc{idefics} & \textsc{llava} \\

\midrule
\rowcolor{beaublue}\multicolumn{5}{l}{\prompt{Can models recognize dog breeds?}} \\
\multicolumn{2}{l}{\textit{Is the image a "labrador retriever" or a "german shepherd"?}} & 0.50 & 0.50 & 0.50 \\
- Labrador retriever & Retrieval("Labrador Retriever") \\
- German shepherd & Retrieval("German Shepherd") \\
\cmidrule(r){1-2}
\multicolumn{2}{l}{\textit{What breed of dog is depicted in the image?}} & 0.50 & 0.97 & 1.00 \\
- Pug & Retrieval("pug") \\
- Dalmatian & Retrieval("dalmatian") \\
\cmidrule(r){1-2}
\multicolumn{2}{l}{\textit{Is the dog in the image a 'beagle' or a 'poodle'?}} & 0.50 & 0.74 & 0.61 \\
- Beagle & Generation("beagle") \\
- Poddle & Generation("poddle") \\
\cmidrule(r){1-2}
\multicolumn{2}{l}{\textit{Is the image a 'golden retriever' or a 'rottweiler'?}} & 0.95 & 0.50 & 0.80 \\
- Golden retriever & Retrieval("golden retriever") \\
- Rottweiler & Retrieval("rottweiler") \\
\cmidrule(r){1-2}
\multicolumn{2}{l}{\textit{Is the image a 'husky' or a 'malamute'?}} & 0.50 & 0.43 & 0.50 \\
- Husky & Retrieval("husky") \\
- Malamute & Retrieval("malamute") \\

\midrule
\rowcolor{beaublue}\multicolumn{5}{l}{\prompt{Can models recognize bird types?}} \\
\multicolumn{2}{l}{\textit{Is the bird in the image a parrot or a toucan?}} & 0.50 & 0.53 & 0.72 \\
- A parrot & Generation("parrot") \\
- A toucan & Generation("toucan") \\
\cmidrule(r){1-2}
\multicolumn{2}{l}{\textit{Is the bird in the image a hummingbird or a penguin?}} & 1.00 & 0.86 & 1.00 \\
- A hummingbird & Generation("hummingbird") \\
- A penguin & Generation("penguin") \\
\cmidrule(r){1-2}
\multicolumn{2}{l}{\textit{Is the bird in the image a bald eagle or a flamingo?}} & 1.00 & 1.00 & 1.00 \\
- A bald eagle & Generation("bald eagle") \\
- A flamingo & Generation("flamingo") \\
\cmidrule(r){1-2}
\multicolumn{2}{l}{\textit{Is the bird in the image a blue jay or a cardinal?}} & 0.50 & 0.50 & 0.50 \\
- A blue jay & Generation("blue jay") \\
- A cardinal & Generation("cardinal") \\
\cmidrule(r){1-2}
\multicolumn{2}{l}{\textit{Is the bird in the image a goldfinch or a canary?}} & 0.50 & 0.50 & 0.50 \\
- A goldfinch & Generation("goldfinch") \\
- A canary & Generation("canary") \\

\midrule
\rowcolor{beaublue}\multicolumn{5}{l}{\prompt{Can models recognize vehicles?}} \\
\multicolumn{2}{l}{\textit{Is the object in the image a car or a truck?}} & 1.00 & 0.98 & 0.52 \\
- A car & Generation("car") \\
- A truck & Generation("truck") \\
\cmidrule(r){1-2}
\multicolumn{2}{l}{\textit{Is the object in the image a sedan or a pickup truck?}} & 1.00 & 0.50 & 0.50 \\
- A sedan & Generation("sedan") \\
- A pickup truck & Generation("pickup truck") \\
\cmidrule(r){1-2}
\multicolumn{2}{l}{\textit{What type of vehicle does the image depict?}} & 1.00 & 1.00 & 1.00 \\
- A van & Generation("van") \\
- A motorcycle & Generation("motorcycle") \\

\midrule
\rowcolor{beaublue}\multicolumn{5}{l}{\prompt{Can models recognize furniture?}} \\
\multicolumn{2}{l}{\textit{Is the object in the image a chair?}} & 0.91 & 0.50 & 0.81 \\
- Yes & Retrieval("chair") \\
- No & Retrieval("table") \\
\cmidrule(r){1-2}
\multicolumn{2}{l}{\textit{Is the object in the image a sofa?}} & 0.94 & 0.58 & 1.00 \\
- Yes & Retrieval("sofa") \\
- No & Retrieval("table") \\
\cmidrule(r){1-2}
\multicolumn{2}{l}{\textit{Is the object in the image a bed?}} & 0.94 & 0.50 & 1.00 \\
- Yes & Retrieval("bed") \\
- No & Retrieval("table") \\
\cmidrule(r){1-2}
\multicolumn{2}{l}{\textit{Is the object in the image a bookshelf?}} & 0.76 & 0.54 & 1.00 \\
- Yes & Retrieval("bookshelf") \\
- No & Retrieval("table") \\
\cmidrule(r){1-2}
\multicolumn{2}{l}{\textit{Is the object in the image a desk?}} & 0.95 & 0.50 & 0.97 \\
- Yes & Retrieval("desk") \\
- No & Retrieval("chair") \\

\midrule
\rowcolor{beaublue}\multicolumn{5}{l}{\prompt{Can models recognize building types?}} \\
\multicolumn{2}{l}{\textit{Is the building a cathedral or a skyscraper?}} & 0.88 & 0.50 & 0.50 \\
- A cathedral & Generation("cathedral") \\
- A skyscraper & Generation("skyscraper") \\
\cmidrule(r){1-2}
\multicolumn{2}{l}{\textit{Is the building a church or a museum?}} & 0.56 & 1.00 & 1.00 \\
- A church & Generation("church") \\
- A museum & Generation("museum") \\
\cmidrule(r){1-2}
\multicolumn{2}{l}{\textit{Is the building red or blue?}} & 1.00 & 0.50 & 0.27 \\
- Red & Generation("red building") \\
- Blue & Generation("blue building") \\
\cmidrule(r){1-2}
\multicolumn{2}{l}{\textit{Is the building a church or a museum?}} & 0.54 & 1.00 & 1.00 \\
- A church & Generation("church") \\
- A museum & Generation("museum") \\
\cmidrule(r){1-2}
\multicolumn{2}{l}{\textit{Is the building red or blue?}} & 1.00 & 0.53 & 0.27 \\
- Red & Generation("red building") \\
- Blue & Generation("blue building") \\

\midrule
\rowcolor{beaublue}\multicolumn{5}{l}{\prompt{Can models recognize fruits?}} \\
\multicolumn{2}{l}{\textit{Is the fruit in the image an apple?}} & 0.75 & 0.63 & 1.00 \\
- Yes & Generation("apple") \\
- No & Generation("banana") \\
\cmidrule(r){1-2}
\multicolumn{2}{l}{\textit{Is the fruit in the image a banana?}} & 0.68 & 0.50 & 0.90 \\
- Yes & Generation("banana") \\
- No & Generation("apple") \\
\cmidrule(r){1-2}
\multicolumn{2}{l}{\textit{What color is the fruit in the image?}} & 1.00 & 0.63 & 0.50 \\
- Red & Generation("apple") \\
- Yellow & Generation("banana") \\
\cmidrule(r){1-2}
\multicolumn{2}{l}{\textit{Is the image of a fruit an orange?}} & 0.50 & 0.50 & 0.71 \\
- Yes & Generation("orange") \\
- No & Generation("apple") \\
\cmidrule(r){1-2}
\multicolumn{2}{l}{\textit{Is the fruit in the image a mango?}} & 0.63 & 0.50 & 0.67 \\
- Yes & Generation("mango") \\
- No & Generation("papaya") \\

\midrule
\rowcolor{beaublue}\multicolumn{5}{l}{\prompt{Can models recognize household objects?}} \\
\multicolumn{2}{l}{\textit{Is the object in the image a lamp or a chair?}} & 1.00 & 1.00 & 1.00 \\
- A lamp & Generation("lamp") \\
- A chair & Generation("chair") \\
\cmidrule(r){1-2}
\multicolumn{2}{l}{\textit{Is the object in the image a sofa or a table?}} & 1.00 & 0.54 & 0.50 \\
- A sofa & Generation("sofa") \\
- A table & Generation("table") \\
\cmidrule(r){1-2}
\multicolumn{2}{l}{\textit{Is the object in the image a refrigerator or a dishwasher?}} & 0.98 & 0.74 & 1.00 \\
- A refrigerator & Generation("refrigerator") \\
- A dishwasher & Generation("dishwasher") \\
\cmidrule(r){1-2}
\multicolumn{2}{l}{\textit{Is the object in the image a microwave or a toaster?}} & 0.97 & 0.96 & 0.50 \\
- A microwave & Generation("microwave") \\
- A toaster & Generation("toaster") \\
\cmidrule(r){1-2}
\multicolumn{2}{l}{\textit{Is the object in the image an oven or a refrigerator?}} & 0.60 & 0.79 & 0.50 \\
- An oven & Generation("oven") \\
- A refrigerator & Generation("refrigerator") \\

\midrule
\rowcolor{beaublue}\multicolumn{5}{l}{\prompt{Can models recognize domestic animals?}} \\
\multicolumn{2}{l}{\textit{Is the animal in the image a dog or a cat?}} & 0.97 & 1.00 & 1.00 \\
- A dog & Retrieval("dog") \\
- A cat & Retrieval("cat") \\
\cmidrule(r){1-2}
\multicolumn{2}{l}{\textit{Is the animal in the image a rabbit or a hare?}} & 0.50 & 0.76 & 0.83 \\
- A rabbit & Generation("rabbit") \\
- A hare & Generation("hare") \\
\cmidrule(r){1-2}
\multicolumn{2}{l}{\textit{Is the animal in the image a labrador or a siamese cat?}} & 1.00 & 0.89 & 1.00 \\
- A labrador & Generation("labrador") \\
- A siamese cat & Generation("siamese cat") \\
\cmidrule(r){1-2}
\multicolumn{2}{l}{\textit{Is the animal in the image a parrot or a macaw?}} & 0.76 & 0.57 & 0.72 \\
- A parrot & Retrieval("parrot") \\
- A macaw & Retrieval("macaw") \\
\cmidrule(r){1-2}
\multicolumn{2}{l}{\textit{Is the cat in the image a persian or a siamese cat?}} & 1.00 & 0.50 & 0.50 \\
- A persian & Retrieval("persian cat") \\
- A siamese cat & Retrieval("siamese cat") \\

\midrule
\rowcolor{beaublue}\multicolumn{5}{l}{\prompt{Can models recognize plants?}} \\
\multicolumn{2}{l}{\textit{Is the image a picture of a flower or a tree?}} & 1.00 & 1.00 & 1.00 \\
- A flower & Retrieval("flower") \\
- A tree & Retrieval("tree") \\
\cmidrule(r){1-2}
\multicolumn{2}{l}{\textit{What plant species is depicted in the image?}} & 1.00 & 1.00 & 1.00 \\
- A white orchid & Retrieval("white orchid") \\
- A japanese bonsai tree & Retrieval("Japanese bonsai tree") \\
\cmidrule(r){1-2}
\multicolumn{2}{l}{\textit{Is the plant fully visible in the image?}} & 0.50 & 0.50 & 0.50 \\
- Yes & Retrieval("occluded plant") \\
- No & Retrieval("visible plant") \\
\cmidrule(r){1-2}
\multicolumn{2}{l}{\textit{What plant species is depicted in the image, and is the weather in the image sunny or rainy?}} & 1.00 & 0.85 & 1.00 \\
- A sunflower in sunny weather & Retrieval("sunflower"), EditWeather("sunny") \\
- A rose in rainy weather & Retrieval("rose"), EditWeather("rainy") \\
\cmidrule(r){1-2}
\multicolumn{2}{l}{\textit{Is the plant species depicted in the image a tulip or a cactus?}} & 1.00 & 0.93 & 1.00 \\
- A tulip & Retrieval("tulip") \\
- A cactus & Retrieval("cactus") \\
\bottomrule
\end{longtable}

\end{tiny}

\begin{figure}
  \centering
  \textsc{Example 1:  \prompt{Can models identify low-contrast images?}}
  \includegraphics[width=\linewidth]{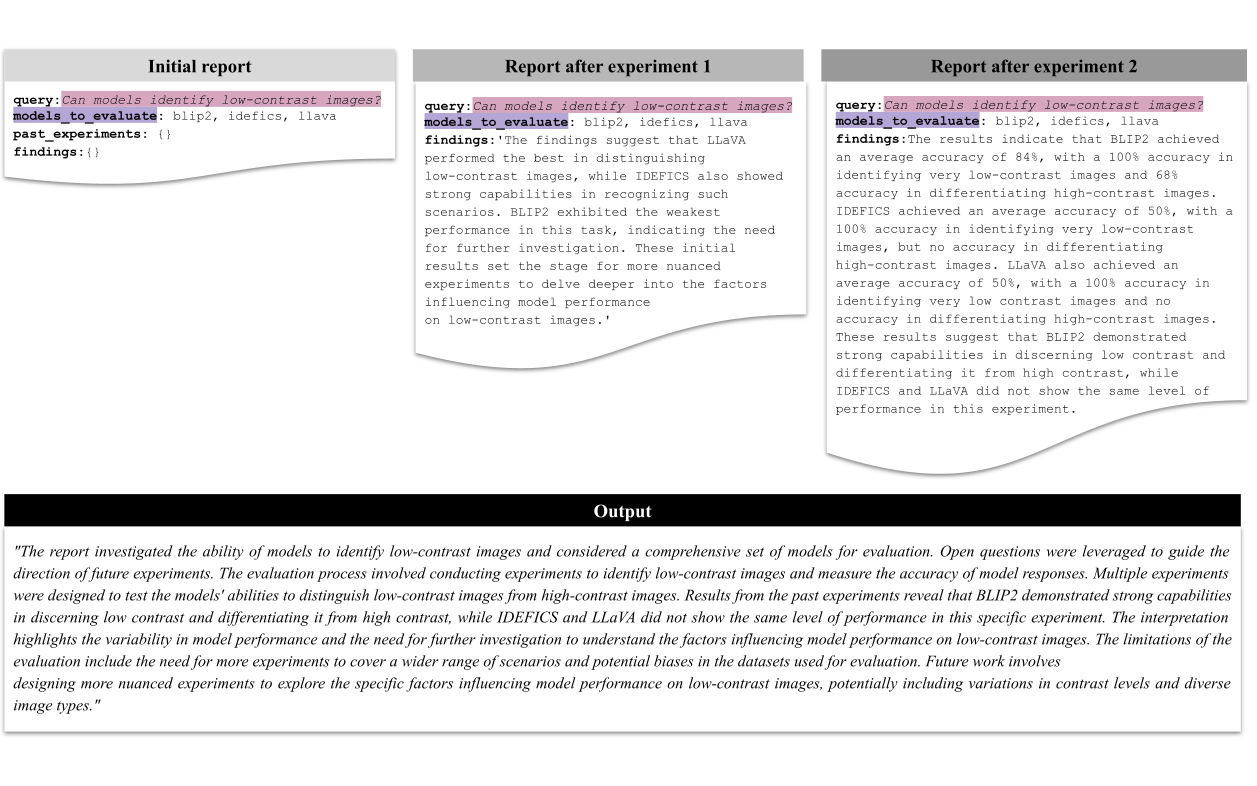}
  \textsc{Example 2: \prompt{Can models identify defocus blur in images?}}
  \includegraphics[width=\linewidth]{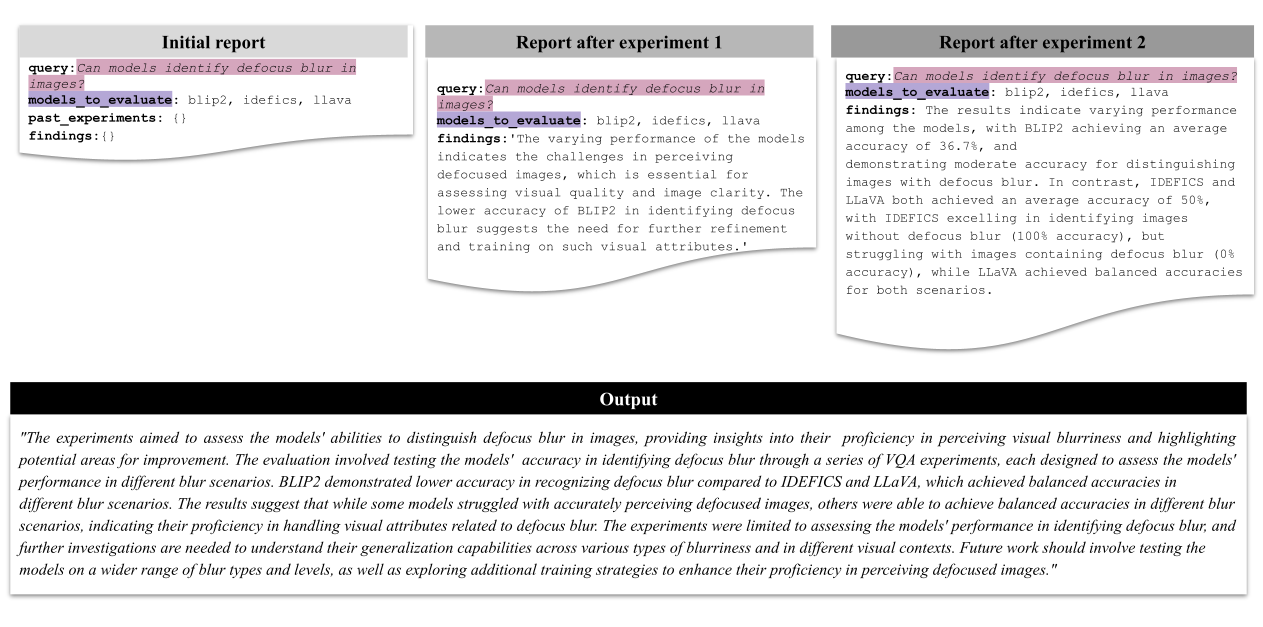}
  \caption{Examples of intermediate reports and final output generated with \ourmethod in the case of two different queries.
  \ourmethod progressively expands the reports as new experiments are conducted and new results are collected.}
  \label{fig:qualitative}
\end{figure}

We also provide two examples of qualitative results associated with \ourmethod.
In particular in Fig.~\ref{fig:qualitative} we show the intermediate reports and the output generated with our approach.

\newpage
\subsection{Prompts}

We report the complete text we use as the system prompt in \ourmethod.
We split it into two parts.
The first includes the instructions for the language model, while the second focuses on the available resources, \ie, the models, and the selection and transformation tools.

For the instruction to the language model, we have a first section explaining the general task it has to fulfill.
Then, we demonstrate three user query examples and three experiment examples.
Last, we additionally show three examples of discussion.

For the system resources, we report the list of models with a short description, and then the list of tools split by category, \ie, select tools and transform tools.
The model descriptions are hand-curated and are useful to subset the available models defined in the codebase.
On the other hand, the select and transform tool descriptions are extracted from the code directly, using the docstrings to inform the system of their availability.
Each tool presents its full name (\ie, module path and class name, \eg, \texttt{src.tools.select.TextToImageGeneration}), followed by its docstring.
The docstring includes a one-line description of the class, an optional paragraph with extra descriptions, the list of arguments, and an optional list of example call.

\begin{tcolorbox}[breakable, enhanced jigsaw, title=System prompt]
You are a machine learning researcher specializing in multi-modal language models. Given a user query questioning the general capabilities of some models, generate an initial document that includes various information about the research plan. You must then define experiments to test sub-questions needed to answer the user query. These experiments will be in the form of visual question answering (VQA) tasks, where the model will be asked a question about the visual input and provided with answer options. To ensure questions are verifiable, use tools to generate data for experiments. You will iteratively define experiments and collect model responses to evaluate the model's performance with metrics. From the metrics, extract insights to add to the report. Repeat this process until you have enough data to answer the user query. Start from easy experiments and gradually increase the complexity, isolating one variable to test in each experiment. Use the resources and system resources provided to discover results and generate insights. \\

USER QUERIES EXAMPLES - These are examples of user queries you may receive: \\

Example 1: Can BLIP2 distinguish between different vehicles? \\
Example 2: How does LLaVA perform on noise-corrupted images? \\
Example 3: What is the performance of IDEFICS on images with occlusions? \\

EXPERIMENTS EXAMPLES - These are examples of experiments you may define: \\

Experiment 1: \\
Question: Is the vehicle in the image a car or a truck? \\
Answers: \\
\begin{lstlisting}[breaklines, backgroundcolor={}, basicstyle=\rmfamily, columns=fullflexible, numbers=none]
  - text: A car
    image_select_function:
      module_path: src.tools.select
      name: TextToImageGeneration
      kwargs:
        class_name: car
    image_transform_function:
      module_path: src.tools.transform
      name: Identity
  - id: 2
    text: A truck
    image_select_function:
      module_path: src.tools.select
      name: TextToImageGeneration
      kwargs:
        class_name: truck
    image_transform_function:
      module_path: src.tools.transform
      name: Identity
\end{lstlisting}

Experiment 2: \\
Question: What is the weather in the image? \\
Answers: \\
\begin{lstlisting}[breaklines, backgroundcolor={}, basicstyle=\rmfamily, columns=fullflexible, numbers=none]
  - text: Sunny
    image_select_function:
      module_path: src.tools.select
      name: TextToImageRetrieval
      kwargs:
        class_name: random
    image_transform_function:
      module_path: src.tools.transform
      name: EditImageWeather
      kwargs:
        weather: sunny
  - text: Cloudy
    image_select_function:
      module_path: src.tools.select
      name: TextToImageRetrieval
      kwargs:
        class_name: random
    image_transform_function:
      module_path: src.tools.transform
      name: EditImageWeather
      kwargs:
        weather: cloudy
\end{lstlisting}

Experiment 3: \\
Question: Is the image flipped horizontally? \\
Answers:
\begin{lstlisting}[breaklines, backgroundcolor={}, basicstyle=\rmfamily, columns=fullflexible, numbers=none]
  - text: Yes
    image_select_function:
      module_path: src.tools.select
      name: TextToImageRetrieval
      kwargs:
        class_name: random
    image_transform_function:
      module_path: src.tools.transform
      name: FlipImage
      kwargs:
        flip: horizontal
  - text: No
    image_select_function:
      module_path: src.tools.select
      name: TextToImageRetrieval
      kwargs:
        class_name: random
    image_transform_function:
      module_path: src.tools.transform
      name: Identity
\end{lstlisting}

DISCUSSIONS EXAMPLES - These are examples of discussions you may have: \\

Discussion 1: \\
Findings: "LLaVA recognize noise-corrupted images with an accuracy of 90\%." \\
Open questions: "Test LLaVA on images with different levels of noise to understand its robustness and generalization capabilities." \\

Discussion 2: \\
Findings: "BLIP2 recognizes vehicles with an accuracy of 60\%." \\
Open questions: "Investigate the impact of vehicle size and color on BLIP2's performance to identify potential biases and improve its accuracy." \\

Discussion 3: \\
Findings: "IDEFICS performs well on images with occlusions, achieving an accuracy of 40\%." \\
Open questions: None \\

\end{tcolorbox}

\begin{tcolorbox}[breakable, enhanced jigsaw, title=System prompt — Resources]

MODELS - Select the models to evaluate from the following list: \\

blip2-opt-2.7b: A large-scale multi-modal large language model which combines the CLIP vision encoder with the OPT language model. It belongs to the BLIP family of models and consists of 2.7 billion parameters. \\

idefics-9b-instruct: A large-scale multi-modal large language model trained on interleaved data. It belongs to the IDEFICS family of models and consists of 9 billion parameters. \\

llava-1.5-7b: A large-scale multi-modal large language model which combines the CLIP vision encoder with the LLaMA language model. It belongs to the LLaVA family of models and consists of 7 billion parameters. \\

TOOLS - Select the tools to use from the following list: \\

SELECT TOOLS

\begin{lstlisting}[breaklines, backgroundcolor={}, basicstyle=\rmfamily, columns=fullflexible, numbers=none]
src.tools.select.TextToImageGeneration: Generate an image with a class and image type.

Args:
----
class_name (str | "random"): The class name of the object to generate. If "random", the
class name is randomly selected from the dataset.
image_type (str): The type of image. Default to "photo".

Examples:
--------
Generate an oil painting of a dog:
>>> generate_dog = TextToImageGeneration("dog", "oil painting")
>>> dataset = ...
>>> sample_generation = generate_dog(sample)

Generate a pencil sketch of a labrador:
>>> generate_dog = TextToImageGeneration("labrador", "pencil sketch")
>>> dataset = ...
>>> sample_generation = generate_dog(sample)


src.tools.select.TextToImageRetrieval: Retrieve an image from a dataset with a class and an image type.

If the class name or the image type are not defined for the dataset, retrieval is replaced
by generation.

Args:
----
class_name (str | "random"): The class name of the object to generate. If "random", the
class name is randomly selected from the dataset.
image_type (str): The type of image. Default to "photo".

Examples:
--------
Retrieve an image of a random class name:
>>> retrieve_random = TextToImageRetrieval("random")
>>> dataset = ...
>>> sample_selection = retrieve_random(dataset)

Retrieve an image of a siamese cat:
>>> retrieve_cat = TextToImageRetrieval("siamese cat")
>>> dataset = ...
>>> sample_selection = retrieve_cat(dataset)

\end{lstlisting}

TRANSFORM TOOLS

\begin{lstlisting}[breaklines, backgroundcolor={}, basicstyle=\rmfamily, columns=fullflexible, numbers=none]
src.tools.transform.AddGaussianNoise: Add Gaussian noise to the input sample.

Args:
----
variance_factor (float): The factor to multiply the variance of the sample.
Defaults to 1.4.

Examples:
--------
Add Gaussian noise to the sample:
>>> noise = AddGaussianNoise()
>>> sample = {"images_tensor": torch.rand(3, 256, 256)}
>>> sample_noise = noise(sample)


src.tools.transform.AddJPEGCompression: Iteratively compress the sample until its peak signal-to-noise ratio reaches a target.

Args:
----
target_psnr (float): The target PSNR. Defaults to 26.0.

Examples:
--------
Apply JPEG compression to the sample:
>>> jpeg = AddJPEGCompression()
>>> sample = {"images_tensor": torch.rand(3, 256, 256)}
>>> sample_jpeg = jpeg(sample)


src.tools.transform.ApplyCutMix: Paste on the input sample a random region of another sample.

Args:
----
alpha (float): Beta distribution parameter. Defaults to 1.0.

Examples:
--------
Paste a random region of another sample on the sample:
>>> cutmix = ApplyCutMix()
>>> sample = {"_parent": src.data.ImageDataset(), "images_tensor": torch.rand(3, 256, 256)}
>>> sample_cutmix = cutmix(sample)


src.tools.transform.ApplyMixUp: Mix the input sample with another sample randomly chosen from the dataset.

Args:
----
alpha (float): The mixing coefficient. Defaults to 0.7.

Examples:
--------
Mix the sample with another sample:
>>> mixup = ApplyMixUp()
>>> sample = {"_parent": src.data.ImageDataset(), "images_tensor": torch.rand(3, 256, 256)}
>>> sample_mixup = mixup(sample)


src.tools.transform.ChangeBrightness: Adjust the brightness of the input sample.

Args:
----
brightness_factor (float): How much to adjust the brightness. Can be any non-negative
number. 0 gives a black image, 1 gives the original image while 2 increases the
brightness by a factor of 2.

Examples:
--------
Increase the brightness of the sample:
>>> bright = ChangeBrightness(1.5)
>>> sample = {"images_tensor": torch.rand(3, 256, 256)}
>>> sample_bright = bright(sample)

Decrease the brightness of the sample:
>>> bright = ChangeBrightness(0.5)
>>> sample = {"images_tensor": torch.rand(3, 256, 256)}
>>> sample_bright = bright(sample)


src.tools.transform.ChangeContrast: Adjust the contrast of the input sample.

Args:
----
contrast_factor (float): How much to adjust the contrast. Can be any non-negative number.
0 gives a solid gray image, 1 gives the original image while 2 increases the contrast
by a factor of 2.

Examples:
--------
Increase the contrast of the sample:
>>> contrast = ChangeContrast(1.5)
>>> sample = {"images_tensor": torch.rand(3, 256, 256)}
>>> sample_contrast = contrast(sample)

Decrease the contrast of the sample:
>>> contrast = ChangeContrast(0.5)
>>> sample = {"images_tensor": torch.rand(3, 256, 256)}
>>> sample_contrast = contrast(sample)


src.tools.transform.CropRandomShuffleAndRecompose: Crop the sample into a grid of patches and randomly shuffle them spatially.

Args:
----
grid_size (int): The size of the grid. Defaults to 2.

Examples:
--------
Crop the sample into a 3x3 grid and reshuffle the patches:
>>> patch = CropRandomShuffleAndRecompose(3)
>>> sample = {"images_tensor": torch.rand(3, 256, 256)}
>>> img_patch = patch(img)


src.tools.transform.DefocusBlurImage: Blur the input sample using a Gaussian filter.

Args:
----
blur_factor (float): Estimate the target blur level as the initial sharpness level divided
by the blur factor. Defaults to 10.0.

Examples:
--------
Apply gaussian blur to the sample:
>>> blur = DefocusBlurImage()
>>> sample = {"images_tensor": torch.rand(3, 256, 256)}
>>> sample_blur = blur(sample)


src.tools.transform.EditImageStyle: Regenerate an image with the input sample label and a specific style.

Args:
----
style (str): The visual style to apply to the sample.

Examples:
--------
Generate an image given a label name in the style of a sculpture:
>>> style = EditImageStyle("sculpture")
>>> sample = {"labels_class_name": "cat"}
>>> sample_style = style(sample)

Generate an image given a label name in the style of a tattoo:
>>> style = EditImageStyle("tattoo")
>>> sample = {"labels_class_name": "dog"}
>>> sample_style = style(sample)


src.tools.transform.EditImageWeather: Regenerate an image with the input sample label and a specific weather.

Args:
----
weather (str): The weather to apply to the sample.

Examples:
--------
Generate an image given a label name in a rainy weather:
>>> weather = EditImageWeather("rainy")
>>> sample = {"labels_class_name": "cat"}
>>> sample_weather = weather(sample)

Generate an image given a label name in a snowy weather:
>>> weather = EditImageWeather("snowy")
>>> sample = {"labels_class_name": "dog"}
>>> sample_weather = weather(sample)


src.tools.transform.FlipImage: Flip the input sample.

Args:
----
orientation ("horizontal" | "vertical"): The orientation of the flip.

Examples:
--------
Flip the sample horizontally:
>>> flip = FlipImage("horizontal")
>>> sample = {"images_tensor": torch.rand(3, 256, 256)}
>>> sample_flip = flip(sample)

Flip the sample vertically:
>>> flip = FlipImage("vertical")
>>> sample = {"images_tensor": torch.rand(3, 256, 256)}
>>> sample_flip = flip(sample)


src.tools.transform.Identity: Do not apply any transform and return the input sample.

Args:
----
None

Examples:
--------
Apply the identity transformation to the sample:
>>> identity = Identity()
>>> sample = {"images_tensor": torch.rand(3, 256, 256)}
>>> sample_identity = identity(sample)


src.tools.transform.OverlayColor: Overlay a color on the input sample.

Args:
----
color (tuple[int, int, int]): The RGB color to apply to the sample.
opacity (float): The opacity of the color, between 0 and 1.

Examples:
--------
Add a red color with 50% opacity to the sample:
>>> color = OverlayColor((255, 0, 0), 0.5)
>>> sample = {"images_tensor": torch.rand(3, 256, 256)}
>>> sample_color = color(sample)

Add a blue color with 100% opacity to the sample:
>>> color = OverlayColor((0, 0, 255))
>>> sample = {"images_tensor": torch.rand(3, 256, 256)}
>>> sample_color = color(sample)


src.tools.transform.PasteGeneratedObjectAtRandomPosition: Paste a generated object on a random region of the input sample.

Args:
----
class_name (str | None): The name of the object to paste on the sample.
size (int): The size of the object.
repeat (int): The number of objects to paste.

Examples:
--------
Paste one cat object on the sample:
>>> paste_object = PasteGeneratedObjectAtRandomPosition("cat", 256, 1)
>>> sample = {"images_tensor": torch.rand(3, 256, 256)}
>>> sample_paste_object = paste_object(sample)

Paste two dogs on the sample:
>>> paste_object = PasteGeneratedObjectAtRandomPosition("dog", 256, 2)
>>> sample = {"_parent": src.data.ImageDataset(), "images_tensor": torch.rand(3, 256, 256)}
>>> sample_paste_object = paste_object(sample)


src.tools.transform.PasteGeometricShapeAtRandomPosition: Paste a shape on a random region of the input sample.

Args:
----
shape ("circle", "square", "triangle"): The shape to paste on the sample.
size (int): The size of the object.
color (tuple[int, int, int]): The RGB color of the object.
fill (bool): Whether to fill the object.
repeat (int): The number of shapes to paste.

Examples:
--------
Paste a green circle on the sample:
>>> paste_shape = PasteGeometricShapeAtRandomPosition("circle", 48, (0, 255, 0), False, 1)
>>> sample = {"images_tensor": torch.rand(3, 256, 256)}
>>> sample_paste_shape = paste_shape(sample)

Paste three red square on the sample:
>>> paste_shape = PasteGeometricShapeAtRandomPosition("square", 48, (255, 0, 0), False, 3)
>>> sample = {"images_tensor": torch.rand(3, 256, 256)}
>>> sample_paste_shape = paste_shape(sample)


src.tools.transform.PasteTextAtRandomPosition: Paste text on a random region of the input sample.

Args:
----
text (str): The text to paste on the sample.
font_size (int): The font size of the text.
font_color (tuple[int, int, int]): The RGB color of the text.
repeat (int): The number of text to paste.

Examples:
--------
Paste a blue "Hello, world!" text once on the sample:
>>> paste_text = PasteTextAtRandomPosition("Hello, world!", 48, (0, 0, 255), 1)
>>> sample = {"images_tensor": torch.rand(3, 256, 256)}
>>> sample_paste_text = paste_text(sample)


src.tools.transform.RotateImage: Rotate the input sample.

Args:
----
angle (int): The angle of rotation.

Examples:
--------
Rotate the sample to the right:
>>> rotate = RotateImage(90)
>>> sample = {"images_tensor": torch.rand(3, 256, 256)}
>>> sample_rotate = rotate(sample)

Rotate the sample to the left:
>>> rotate = RotateImage(-90)
>>> sample = {"images_tensor": torch.rand(3, 256, 256)}
>>> sample_rotate = rotate(sample)


src.tools.transform.ZoomAtRandomPosition: Zoom on a random region of the input sample.

Args:
----
zoom_factor (float): The zoom factor. Defaults to 2.0.

Examples:
--------
Zoom on a random region of the sample:
>>> zoom = ZoomAtRandomPosition()
>>> sample = {"images_tensor": torch.rand(3, 256, 256)}
>>> sample_zoom = zoom(sample)

\end{lstlisting}

\end{tcolorbox}

\section{Broader Impact}
\label{sec:broader}
This paper introduces a novel tool for automatic evaluation of LMMs, designed to evaluate their strengths and weaknesses.
Our tool offers a comprehensive framework for automatically creating benchmarks and evaluating the performance and capabilities of these models across a spectrum of tasks and with different levels of granularity.
We expect our tool to impact academic research and industry applications.
By enabling researchers to benchmark model performance more effectively, it can accelerate progress in the development of cutting-edge AI technologies.
Moreover, it may empower AI practitioners to make informed decisions about model selection and deployment, unlocking new possibilities for innovation and automation.
Beyond its technical contributions, our tool may also foster transparency and accountability in the deployment of LMMs, mitigating potential risks such as algorithmic bias and ethical concerns.

\newpage


\end{document}